%% file: main.tex
\documentclass[10pt,twocolumn,letterpaper]{article}

\usepackage[pagenumbers]{cvpr}

\usepackage[accsupp]{axessibility}

\usepackage{algorithm}
\usepackage{algpseudocode}
\usepackage{booktabs}
\usepackage{pifont}
\usepackage{multirow}
\usepackage{makecell}
\usepackage{graphicx}
\usepackage[table]{xcolor}
\usepackage{array}
\usepackage{textcomp}
\usepackage{paralist}

\input{preamble}

\definecolor{cvprblue}{rgb}{0.21,0.49,0.74}
\usepackage[pagebackref,breaklinks,colorlinks,allcolors=cvprblue]{hyperref}

\def\paperID{32944}
\def\confName{CVPR}
\def\confYear{2026}

\title{RobotSeg: A Model and Dataset for Segmenting Robots in Image and Video}

\author{
\begin{tabular}{c}
Haiyang Mei \quad
Qiming Huang \quad
Hai Ci \quad
Mike Zheng Shou*
\\[6pt]
Show Lab, National University of Singapore
\\[5pt]
{\small\url{https://github.com/showlab/RobotSeg}}
\end{tabular}
}

\begin{document}

\twocolumn[{
\renewcommand\twocolumn[1][]{#1}
\maketitle
\begin{center}
    \vspace{-6mm}
	\centering
	\captionsetup{type=figure}
	\includegraphics[width=.96\textwidth,height=8.9cm]{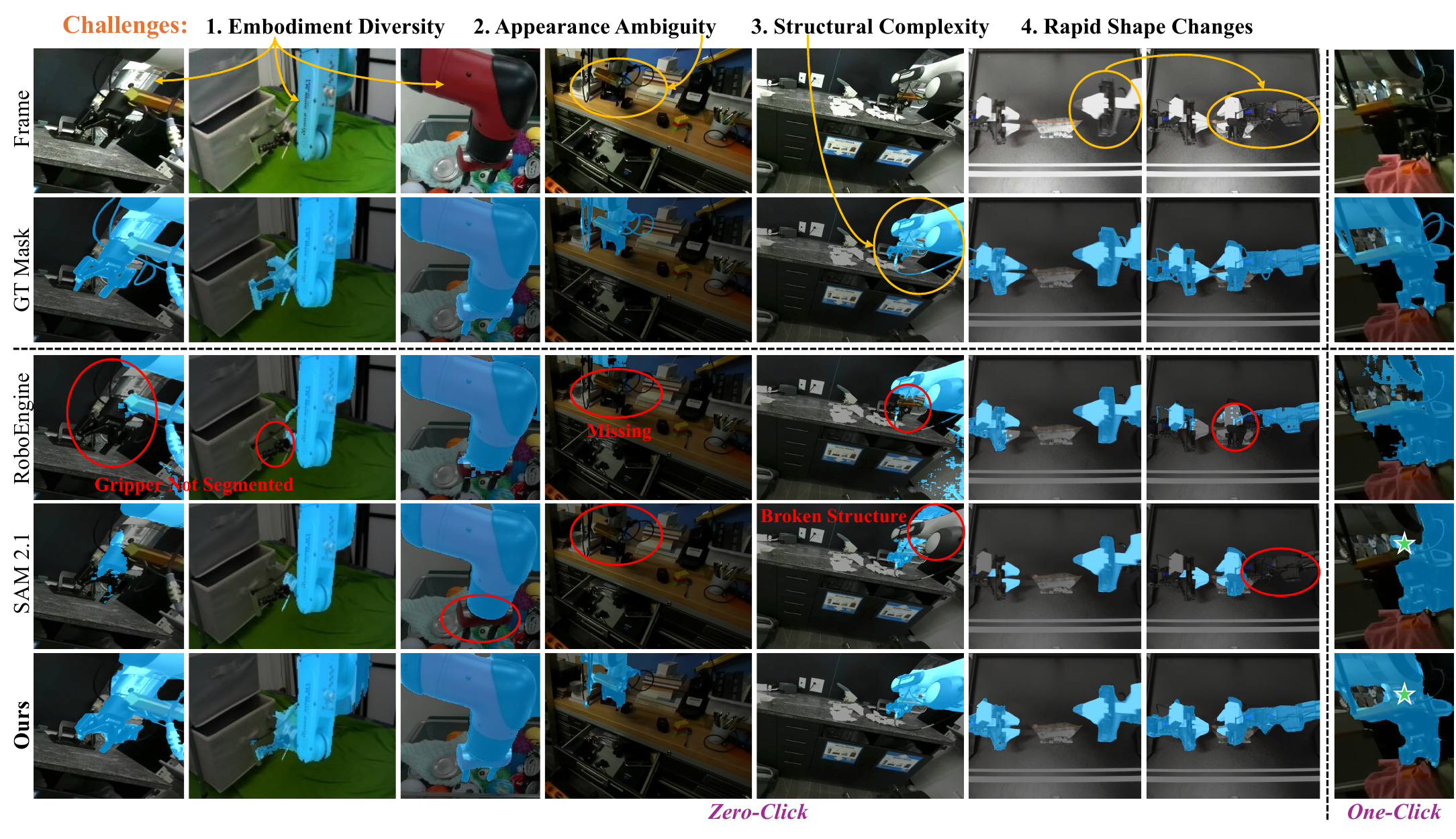}\vspace{-4mm}
    \captionof{figure}{Although existing state-of-the-art segmentation models (\eg, RoboEngine \cite{yuan2025roboengine} and SAM~2.1 \cite{sam2}) are highly capable, surprisingly they struggle to segment robots: they fail to cope with diverse embodiments (1-3 columns), often confuse robots with cluttered backgrounds (4th column), break down when facing complex structures (5th column), and fail to handle rapid shape changes (6th-7th columns). In contrast, our RobotSeg model (last row) achieves robust robot segmentation across diverse embodiments and scenes, and further supports user-provided prompts (\eg, clicks or bounding boxes) to refine the segmentation results (last column).}\vspace{0.8mm}
	\label{fig:teaser}
\end{center}
}]

\renewcommand{\thefootnote}{}
\footnotetext{* Corresponding Author}

\input{sec/0_abstract}
\input{sec/1_introduction}
\input{sec/2_related_work}
\input{sec/3_dataset}
\input{sec/4_methodology}
\input{sec/5_experiments}
\input{sec/6_conclusion}

{
\small
\bibliographystyle{ieeenat_fullname}
\bibliography{main}
}

\clearpage
\input{sec/7_supp}

\end{document}

%% file: preamble.tex
\newcommand{\cmark}{\ding{51}}

%% file: sec/0_abstract.tex
\begin{abstract}
Accurate robot segmentation is a fundamental capability for robotic perception. It enables precise visual servoing for VLA systems, scalable robot-centric data augmentation, accurate real-to-sim transfer, and reliable safety monitoring in dynamic human-robot environments. Despite the strong capabilities of modern segmentation models, surprisingly it remains challenging to segment robots. This is due to robot embodiment diversity, appearance ambiguity, structural complexity, and rapid shape changes. Embracing these challenges, we introduce RobotSeg, a foundation model for robot segmentation in image and video. RobotSeg is built upon the versatile SAM 2 foundation model but addresses its three limitations for robot segmentation, namely the lack of adaptation to articulated robots, reliance on manual prompts, and the need for per-frame training mask annotations, by introducing a structure-enhanced memory associator, a robot prompt generator, and a label-efficient training strategy. These innovations collectively enable a structure-aware, automatic, and label-efficient solution. We further construct the video robot segmentation (VRS) dataset comprising over 2.8k videos (138k frames) with diverse robot embodiments and environments. Extensive experiments demonstrate that RobotSeg achieves state-of-the-art performance on both images and videos, establishing a strong foundation for future advances in robot perception.
\end{abstract}

%% file: sec/1_introduction.tex
\section{Introduction}
\label{sec:introduction}

Robots are increasingly deployed in diverse real-world environments. As robots execute tasks in dynamic and unstructured scenes, accurate robot segmentation becomes a fundamental capability. It enables a wide range of downstream applications, including (i) visual servoing for vision-language-action (VLA) systems by providing precise robot masks that help the policy localize effectors and execute fine-grained control \cite{li2025coa}, (ii) scalable data augmentation by replacing the robot arm with another embodiment to improve generalization across robot types \cite{chen2025rovi,yuan2025roboengine,lin2025data}, (iii) real-to-sim transfer by generating cleaner robot geometry and appearance for building more accurate simulation assets \cite{dan2025x,torne2024reconciling,wang2025recipe}, and (iv) safety monitoring by continuously tracking the robot from a third-person view to detect unsafe motions and avoid collisions with humans \cite{rosenstrauch2018safe,kang2019safety,lin2025digital}. These applications place strong demands on robot segmentation quality across varied embodiments and scenarios, and highlight the need for a robust solution for segmenting robots in image and video.

Despite the strong capabilities of modern segmentation models, surprisingly it remains challenging to segment robots. As illustrated in Figure~\ref{fig:teaser}, state-of-the-art methods (\eg, RoboEngine~\cite{yuan2025roboengine} and SAM~2.1~\cite{sam2}) struggle for several distinctive reasons: (i) robots exhibit highly diverse embodiments across platforms such as Franka, Fanuc Mate, and Sawyer, resulting in substantial variation in shape, size, and appearance (1-3 columns); (ii) their visual textures and colors often resemble cluttered backgrounds, causing models to fail to separate the robot regions from the background (4th column); (iii) robots contain complex articulated structures that current methods frequently segment incompletely, leading to broken or fragmented robot parts (5th column); and (iv) robots undergo rapid shape changes during manipulation, producing highly unstable visual patterns that challenge existing segmentation models (6th-7th columns).
Together, these factors make robot segmentation fundamentally more challenging, underscoring the importance of a dedicated model that works reliably on image and video.

As the state-of-the-art segmentation foundation model, SAM~2~\cite{sam2} nevertheless suffers from method-level deficiencies when applied to robotic perception: (i) it lacks mechanisms for modeling articulated robot structures, often resulting in incomplete or structurally broken masks; (ii) it relies on manual prompts to initiate segmentation, which undermines autonomy and scalability in real-world robotic applications; and (iii) it requires per-frame mask annotation to supervise video training, which is costly and impractical for large-scale robotic datasets.

Motivated by these method-level limitations in robotic perception, we introduce RobotSeg, a structure-aware, automatic, and training-label-efficient foundation model for robot segmentation in image and video. Our design builds upon SAM~2~\cite{sam2} and introduces three key innovations: (1) a structure-enhanced memory associator that integrates temporal context and structural information to enhance feature representations for temporally consistent segmentation of articulated robots; (2) a robot prompt generator that learns semantic priors and historical object cues as prompts for autonomous segmentation; and (3) a label-efficient training strategy that employs forward-backward cycle consistency, semantic consistency, and patch consistency losses, enabling supervision using only the first-frame mask. Furthermore, we construct the video robot segmentation (VRS) dataset, comprising 2,812 videos (138,707 frames) and covering various robot parts (\ie, arm, gripper, and whole robot), diverse embodiments (\eg, Franka, UR5, and WindowX), and a wide range of environments, to enable comprehensive benchmarking and method development. Extensive experiments demonstrate that RobotSeg achieves state-of-the-art performance on both images and videos, establishing a strong foundation for future advances in robot perception.
In summary, our contributions are:
\begin{compactenum}
\item We introduce RobotSeg, the first foundation model for robot segmentation that supports both images and videos, enables fine-grained segmentation of the robot arm, gripper, and whole robot, and offers promptable capabilities for flexible editing and annotation.
\item We construct VRS, the first video robot segmentation benchmark comprising over 2.8k videos (138k frames) spanning diverse robot embodiments, scenes, and lighting conditions, to support comprehensive evaluation and future method development.
\item We design a novel framework that integrates structure-aware memory association, robot prompt generation, and a label-efficient training strategy to enable structure-aware, automatic, and label-efficient video robot segmentation.
\end{compactenum}

%% file: sec/2_related_work.tex
\section{Related Works}
\label{related_work}

\begin{figure*}[t]
    \centering
    \includegraphics[width=\textwidth,height=6cm]{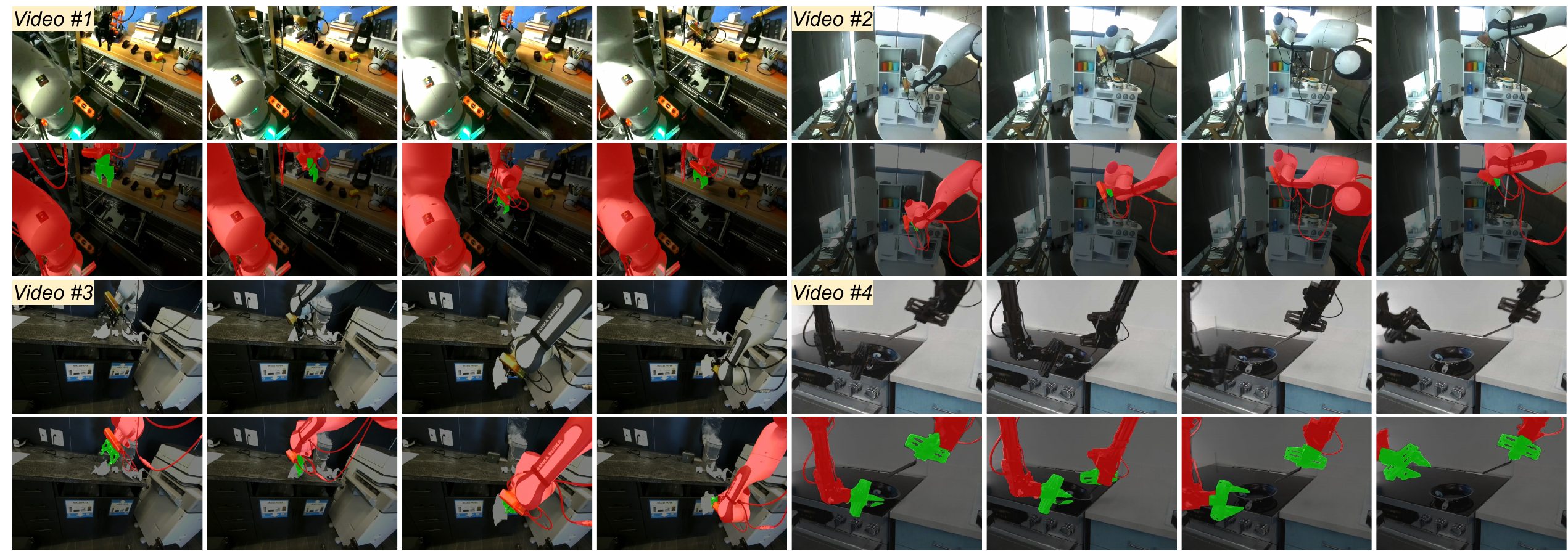}\vspace{-1mm}
    \caption{Examples from our VRS dataset. It encompasses different scenes, lighting conditions, and robot embodiments. Each example shows the RGB sequence (top) and robot annotation masks (bottom), where the robot arm is highlighted in red and the gripper in green.}\vspace{-3mm}
    \label{fig:dataset_example}
\end{figure*}

\noindent\textbf{Robot Segmentation.}
Recently, RoboEngine~\cite{yuan2025roboengine} constructs the first image-level robot segmentation dataset and finetunes a language-conditioned segmentation model, EVF-SAM~\cite{zhang2024evf}, to generate robot masks, which are then fed into Background-Diffusion~\cite{eshratifar2024salient}, a controllable diffusion-based background generator conditioned on masks and scene semantics, to synthesize physics- and task-aware backgrounds, enabling plug-and-play visual augmentation that enhances robot policy adaptability across diverse scenes.
RoVi-Aug~\cite{chen2025rovi} similarly finetunes Segment
Anything Model (SAM) \cite{sam1} using LoRA \cite{hulora} for robot segmentation, and applies a Stable Diffusion model \cite{rombach2022high} conditioned on robot masks via ControlNet \cite{zhang2023adding} for robot-to-robot generation, thereby enhancing cross-embodiment policy transfer.
While these methods demonstrate the effectiveness of segmentation-driven data augmentation, they are restricted to static images and overlook the temporal coherence required for real-world dynamic video perception. Our work addresses this gap by establishing a benchmark and model for segmenting robots in both images and videos.

\noindent\textbf{Semi-Supervised Video Object Segmentation.}
In video object segmentation, the mainstream practice is the semi-supervised setting, where the mask of the target object is provided in the first frame and then propagated throughout the video~\cite{pont20172017}. Early methods achieve this by fine-tuning models on the initial frame~\cite{caelles2017one,maninis2018video,robinson2020learning} or using offline-trained mask propagation networks~\cite{oh2018fast,yang2018efficient,yang2020collaborative}. Subsequent approaches employ recurrent architectures~\cite{xu2018youtube} and attention-based frameworks~\cite{cheng2021rethinking,cheng2022xmem,bekuzarov2023xmem++} to better capture temporal dependencies, while recent advances utilize transformer-based models~\cite{zhang2023joint,wu2023scalable,cheng2024putting} for joint spatial-temporal modeling.
Nevertheless, the reliance on a high-quality mask in the first frame poses a significant limitation for robotics, where manual annotation is often impractical for large-scale or autonomous deployment. This restricts the scalability and real-world applicability of semi-supervised video object segmentation for robot perception.

\noindent\textbf{Language-Conditioned Segmentation.}
An alternative approach for robot segmentation is language-conditioned segmentation, which predicts masks based on textual prompts. Early methods such as CLIPSeg~\cite{luddecke2022image} extend vision-language model CLIP~\cite{radford2021learning} with a lightweight decoder for zero-shot segmentation of arbitrary queries. Recent works leverage large language models (LLMs) to improve mask prediction from complex or implicit queries. LISA~\cite{lai2024lisa} and EVF-SAM~\cite{zhang2024evf} generate segmentation masks using multimodal or early-fusion frameworks, while VideoLISA~\cite{bai2024one} and VISA~\cite{yan2024visa} further introduce temporal modeling and language-guided reasoning to handle challenging video scenarios.
However, these methods either lack temporal modeling or exhibit suboptimal language-to-segmentation transfer, or else rely on large-scale LLMs that introduce significant computational overhead, limiting their effectiveness and deployability in practical robotic applications.

\noindent\textbf{Promptable Visual Segmentation.}
The advent of the Segment Anything Model (SAM)~\cite{sam1} has driven rapid advances in promptable visual segmentation, enabling flexible mask generation for arbitrary objects using input prompts such as points or bounding boxes. Built on the large-scale SA-1B dataset, SAM's versatile prompting mechanism supports zero-shot segmentation and has inspired a range of follow-up models targeting higher segmentation quality~\cite{ke2024segment}, greater efficiency~\cite{xiong2024efficientsam,zhang2023faster,zhang2023mobilesamv2,zhao2023fast,shu2023tinysam,chen20230slimsam}, and broader real-world applications~\cite{ma2024segment,mazurowski2023segment,wu2023medical,chen2024rsprompter,ren2024segment,catsam}. Extending promptable segmentation to the video domain, SAM 2~\cite{sam2} and its subsequent variants~\cite{zhou2025edgetam,ding2024sam2long,Mei_2025_CVPR_I2V,Mei_2026_TPAMI_I2VPP,yang2024samurai,videnovic2024distractor} enable powerful mask propagation across video frames and greatly enhance a wide range of video segmentation applications~\cite{fischer2024sama,dong2024segment,zhu2024medical,shen2024interactive,liu2024surgical,xiong2024sam2,chen2024sam2}.
Despite its versatility, SAM 2 still faces limitations for robotic perception: it relies on manual input (such as clicks or boxes) to initiate segmentation, reducing autonomy in robotic systems, and requires large-scale annotated video masks for training, which are difficult and costly to obtain in diverse robotic scenarios.

%% file: sec/3_dataset.tex
\section{Video Robot Segmentation Dataset}
\label{dataset}

\begin{figure*}[t]
    \centering
    \includegraphics[width=\textwidth]{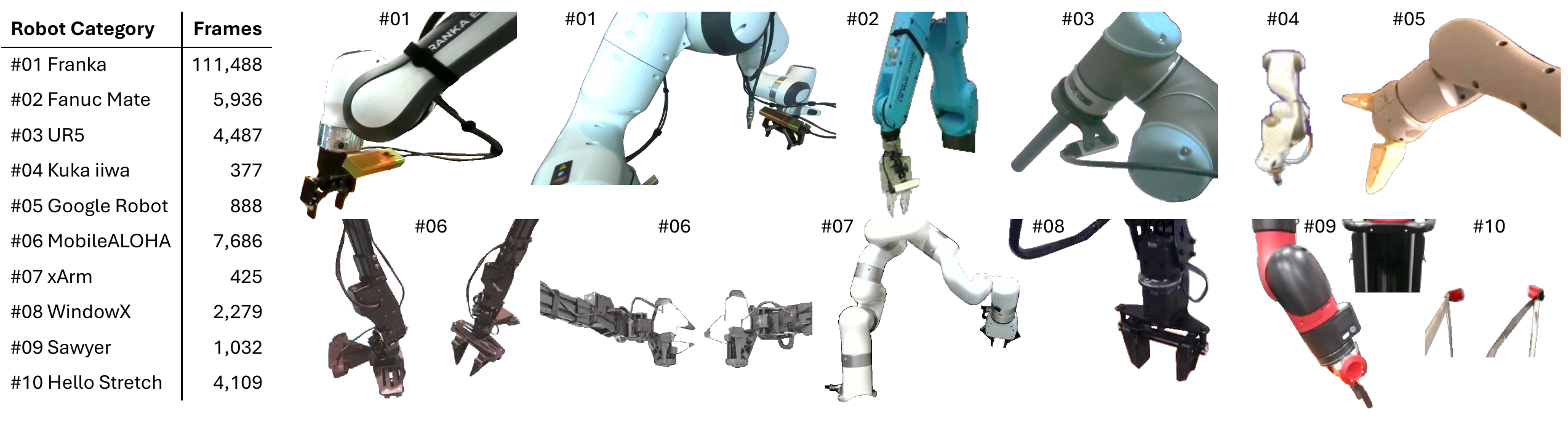}\vspace{-4mm}
    \caption{Robot embodiments in our video robot segmentation (VRS) dataset.}\vspace{-3mm}
    \label{fig:dataset_robot}
\end{figure*}

To train and evaluate video robot segmentation models, we construct the first video robot segmentation dataset, named VRS. As shown in Table~\ref{tab:dataset_stats}, VRS contains 2,812 videos with a total of 138,707 frames, which is 38$\times$ larger than the existing image-level dataset RoboEngine~\cite{yuan2025roboengine} that comprises 3,629 static images. Our VRS covers diverse scenes, robot embodiments, and lighting conditions. Representative video examples and robot embodiments are shown in Figure~\ref{fig:dataset_example} and~\ref{fig:dataset_robot}, respectively.

\textbf{Data Collection.}
To construct VRS, we collect data from DROID (Franka) \cite{khazatsky2024droid,o2024open}, Berkeley Fanuc Manipulation (Fanuc Mate) \cite{zhu2023fanuc}, Columbia PushT (UR5) \cite{chi2025diffusion}, Stanford Kuka Multimodal (Kuka iiwa) \cite{lee2019making}, RoboVQA (Google Robot) \cite{sermanet2024robovqa}, MobileALOHA (MobileALOHA) \cite{fu2025mobile}, UCSD Kitchen (xArm) \cite{ucsd_kitchens}, Berkeley Bridge (WindowX) \cite{walke2023bridgedata}, RoboTurk (Sawyer) \cite{mandlekar2019scaling}, and DobbE (Hello Stretch) \cite{shafiullah2023bringing}. From these datasets, we extract continuous RGB video sequences that capture robot manipulation behaviors. All clips are manually reviewed to remove blurred, duplicated, or incomplete sequences. The retained videos are then annotated by professional annotators following a hierarchical labeling protocol, which not only segments the entire robot but also provides finer-grained masks for the robot arm and gripper, enabling potential applications such as arm/gripper motion analysis and part-aware data augmentation. We randomly split the dataset into 105 testing videos and 2,707 training videos. The testing set is fully annotated, containing 7,203 labeled frames, which is 74× more than the 97 labeled images in RoboEngine-Test~\cite{yuan2025roboengine}. The training set includes 131,504 frames, where only the first frame of each video is annotated. This semi-supervised annotation strategy balances scale, diversity, and practical labeling cost.

\textbf{Dataset Property.}
VRS contributes to the community by providing two key characteristics that distinguish it from existing robot segmentation datasets. First, it is the first video-level benchmark for robot segmentation, offering continuous frame sequences that enable research on mask propagation, temporal modeling, and cross-frame consistency. Second, it is 38$\times$ larger and more diverse than the existing dataset RoboEngine~\cite{yuan2025roboengine}, supporting robust training and comprehensive evaluation, as well as generalization across various robot embodiments, scenes, and lighting conditions. These properties together make VRS a foundation for advancing robot perception in dynamic scenarios.

%% file: sec/4_methodology.tex
\section{Video Robot Segmentation Model}
\label{methodology}

Segmentation has achieved remarkable progress with the emergence of SAM 2 \cite{sam2}, which demonstrates strong capabilities across diverse scenes. However, when applied to robotic perception, several limitations remain: (i) SAM 2 is designed for general objects rather than structured robotic embodiments, potentially leading to inconsistent part segmentation and temporal instability; (ii) it requires manual prompts such as clicks or bounding boxes to initiate segmentation, which limits autonomy and practicality in robotic applications; and (iii) it relies on large-scale video mask annotations for training, which are costly and difficult to obtain for diverse robots and environments. To overcome these challenges, we propose RobotSeg, a specialized video foundation model that enables structure-aware, automatic, and training-label-efficient robot segmentation.

\begin{table}[t]
\centering
\caption{Number statistics of existing dataset and our VRS dataset.}\vspace{-2.5mm}
\label{tab:dataset_stats}
\scriptsize
\renewcommand{\arraystretch}{1.4}
\setlength{\tabcolsep}{1pt}
\begin{tabular}{c|c|c|c|c|c|c|c|c|c}
\hline
\hline
\multirow{2}{*}{\textbf{Dataset}}
 & \multicolumn{3}{c|}{\textbf{Training Set}} 
 & \multicolumn{3}{c|}{\textbf{Testing Set}} 
 & \multicolumn{3}{c}{\textbf{Whole Set}} \\
 \cline{2-10}
 & Video & Frame & Labeled 
 & Video & Frame & Labeled 
 & Video & Frame & Labeled \\
\hline
RoboEngine 
 & 0 & 3,532 & 3,532 
 & 0 & 97 & 97 
 & 0 & 3,629 & 3,629 \\
 \hline
\textbf{VRS (Ours)} 
 & 2,707 & 131,504 & 2,707 
 & 105 & 7,203 & 7,203 
 & 2,812 & 138,707 & 9,910 \\
\hline
\hline
\end{tabular}
\vspace{-5mm}
\end{table}

\begin{figure*}[t]
    \centering
    \includegraphics[width=1\linewidth]{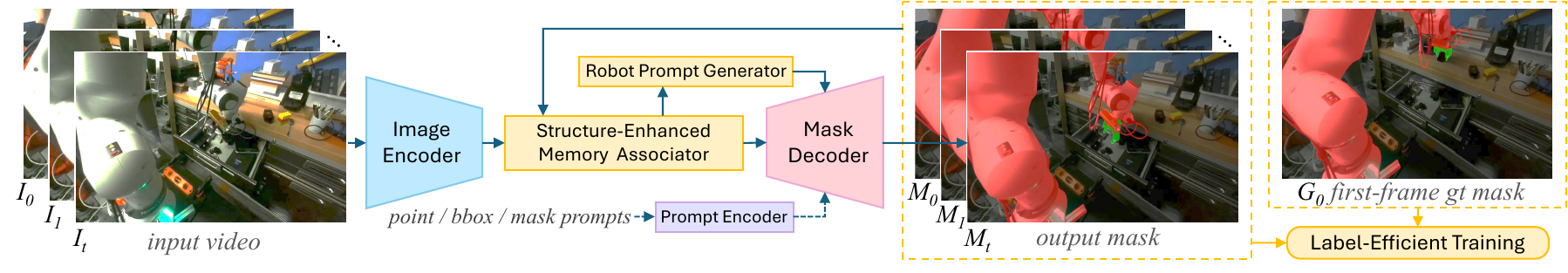}
    \vspace{-0.6cm}
    \caption{Overview of our RobotSeg. Building upon SAM~2~\cite{sam2}, it introduces a structure-enhanced memory associator, a robot prompt generator, and a label-efficient training strategy to enable structure-aware, automatic, and training-label-efficient robot segmentation.}
    \label{fig:pipeline}
    \vspace{-0.25cm}
\end{figure*}

As illustrated in Figure~\ref{fig:pipeline}, RobotSeg takes a video as input, where the backbone extracts frame-wise visual features that are enhanced by the structure-enhanced memory associator SEMA (Section \ref{sec:sema}), which incorporates past frame features and segmentations as memory and performs structure-aware association to inject temporal robot context into the current frame. Building upon the memory, the robot prompt generator RPG (Section \ref{sec:rpg}) produces semantic robot prompts to guide the segmentation process without requiring manual point or box inputs. The resulting robot prompts, together with optional user refinement prompts, guide the mask decoder to generate the robot mask for the current frame. Finally, a label-efficient training strategy (Section \ref{sec:lets}) supervises the model using only the first-frame ground-truth mask, enabling label-efficient video learning.

\subsection{Structure-Enhanced Memory Associator}
\label{sec:sema}
\begin{figure}[t]
    \centering
    \includegraphics[width=1\linewidth]{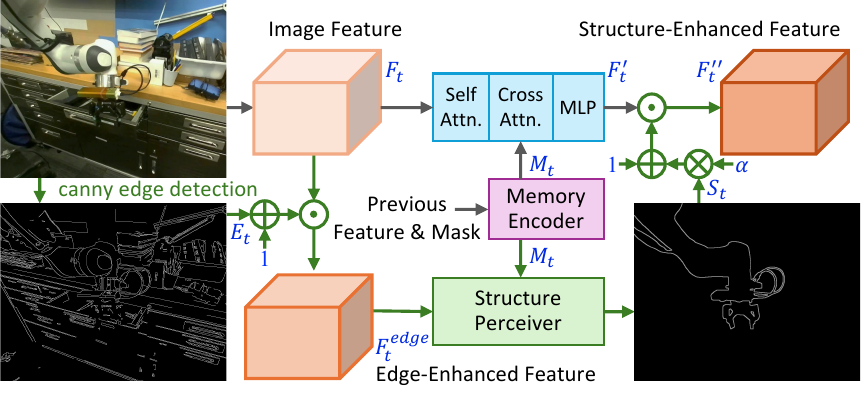}
    \vspace{-0.6cm}
    \caption{Illustration of the structure-enhanced memory associator. It encodes previous features and masks into memory to guide temporal context integration in the top branch and robot boundary perception in the bottom branch, where detected boundaries are used for structure enhancement of the top features.}
    \label{fig:sema}
    \vspace{-0.3cm}
\end{figure}

The SEMA is designed to propagate and associate target information from previous frames to enhance object representations in the current frame features. Considering that robots exhibit inherent geometric regularities and articulated structural priors, SEMA integrates structure-aware modeling into this process. As illustrated in Figure~\ref{fig:sema}, SEMA encodes previous frame features and masks into memory to guide both temporal context integration and structure-aware enhancement of the current frame features.

Formally, SEMA first encodes the features and masks from previous frames into a memory representation \(M_t\), which serves as the reference for temporal association. The current frame features \(F_t\) are then refined through a sequence of self-attention $\mathrm{SelfAttn}$, cross-attention $\mathrm{CrossAttn}$, and feed-forward $\mathrm{MLP}$ operations:
\begin{equation}
F_t' = \mathrm{MLP}\big(\mathrm{CrossAttn}(\mathrm{SelfAttn}(F_t),\, M_t)\big).
\end{equation}
In parallel, the structure branch first extracts an edge map \(E_t = \mathcal{C}(I_t)\) from the current image \(I_t\) using the Canny filter \(\mathcal{C}\), and modulates the current-frame features with \(E_t\):
\begin{equation}
F_t^{\text{edge}} = F_t \odot (1 + E_t),
\end{equation}
where \(\odot\) denotes element-wise multiplication. The edge-enhanced features \(F_t^{\text{edge}}\) are then processed by a structure perceiver, which consists of a multi-scale feature extractor \(\mathcal{MS}(\cdot)\) and a cross-attention module 
for capturing robot-aware structural cues:
\begin{equation}
\begin{gathered}
F_t^{\text{ms}} = \mathcal{MS}(F_t^{\text{edge}}), \\
S_t = \sigma(\mathrm{CrossAttn}(F_t^{\text{ms}},\, M_t)),
\end{gathered}
\end{equation}
where \(\sigma(\cdot)\) denotes the sigmoid function. The resulting structure map \(S_t\) modulates the temporally enhanced features \(F_t'\) to obtain the final structure-enhanced features:
\begin{equation}
F_t'' = F_t' \odot \left(1 + \alpha S_t\right),
\end{equation}
where \(\alpha\) is a learnable modulation weight. The structure map \(S_t\) is supervised during the training.

\subsection{Robot Prompt Generator}
\label{sec:rpg}
The RPG is designed to enable autonomous and temporally consistent robot segmentation without relying on manual prompts such as clicks or bounding boxes. Inspired by the prompt-guided segmentation paradigm in SAM \cite{sam1,sam2}, we propose to generate robot tokens that guide the segmentation process. Specifically, RPG produces two types of tokens: class tokens that provide semantic priors for autonomous perception, and object tokens that convey temporal cues derived from historical observations.

\begin{figure}[t]
    \centering
    \includegraphics[width=1\linewidth]{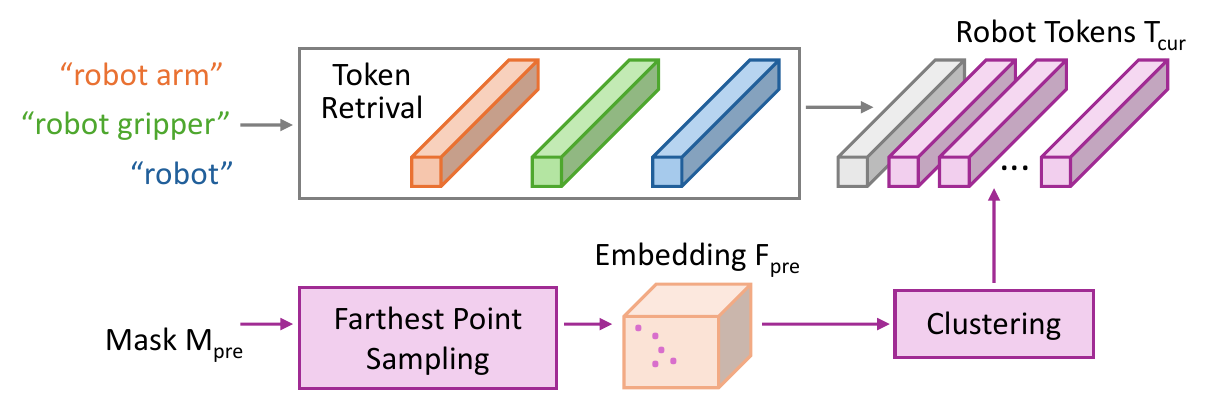}
    \vspace{-0.6cm}
    \caption{Illustration of the robot prompt generator. It generates two types of robot tokens: class tokens retrieved according to the segmentation target (\eg, arm, gripper, or robot), and object tokens derived by clustering historical features within masked regions.}
    \label{fig:rpg}
\end{figure}

As illustrated in Figure~\ref{fig:rpg}, the RPG constructs two complementary types of robot tokens to provide both semantic and temporal guidance. The class tokens are selected from a learnable token bank according to the specified target category (\eg, robot arm, robot gripper, or robot), offering class-level priors to enable autonomous perception. In contrast, the object tokens are dynamically extracted from historical memory by clustering features within the previously segmented regions. To capture both coarse structures and fine-grained details, we adopt a hierarchical clustering strategy (Algorithm \ref{alg:rpg}) that first groups the memory foreground into region-level segments and then further decomposes each region into subclusters. The resulting tokens, together with the class tokens, function as segmentation prompts that guide the following mask decoder in generating the robot mask for the current frame.

\begin{algorithm}[t]
\caption{Hierarchical clustering strategy for object token generation in the robot prompt generator.}
\label{alg:rpg}
\small
\begin{algorithmic}[1]
\Require Previous-frame features $F_{\text{pre}} \in \mathbb{R}^{N \times H \times W \times C}$, previous-frame masks $M_{\text{pre}} \in \{0,1\}^{N \times H \times W}$, number of macro regions $R$, and micro clusters per region $S$
\Ensure Robot tokens for the current frame $T_{\text{cur}} \in \mathbb{R}^{N \times R \times (S \cdot C)}$
\vspace{0.2em}
\For{each memory frame $n = 1$ to $N$}
    \State Extract foreground regions using $M_{\text{pre}}[n]$
    \State Sample $R$ macro centers $\{c_{n,r}\}_{r=1}^R \leftarrow \text{FPS*}(M_{\text{pre}}[n])$
    \State Generate $R$ robot region masks $\{M_{n,r}\}_{r=1}^R \leftarrow \text{KMeans}(F_{\text{pre}}[n],$ initialized by $\{c_{n,r}\}_{r=1}^R$)
    \For{each macro region $r = 1$ to $R$}
        \State Sample $S$ micro centers $\{c_{n,r,s}\}_{s=1}^S \leftarrow \text{FPS*}(M_{n,r})$
        \State Compute $S$ micro prototypes $\{p_{n,r,s}\}_{s=1}^S \leftarrow \text{KMeans}(F_{\text{pre}}[n],$ initialized by $\{c_{n,r,s}\}_{s=1}^S$)
        \State Concatenate: $t_{n,r} = [p_{n,r,1}; \dots; p_{n,r,S}] \in \mathbb{R}^{S \cdot C}$
    \EndFor
    \State Form object token: $T_n = [t_{n,1}; \dots; t_{n,R}] \in \mathbb{R}^{R \times (S \cdot C)}$
\EndFor
\State Stack all memory-frame tokens: $T_{\text{cur}} = [T_1; \dots; T_N] \in \mathbb{R}^{N \times R \times (S \cdot C)}$
\State \Return $T_{\text{cur}}$
\end{algorithmic}
\footnotesize{* FPS denotes Farthest Point Sampling used for spatially diverse initialization within the masked regions.}
\end{algorithm}

\subsection{Label-Efficient Training Strategy}
\label{sec:lets}
\begin{figure*}[t]
    \centering
    \includegraphics[width=.9\linewidth]{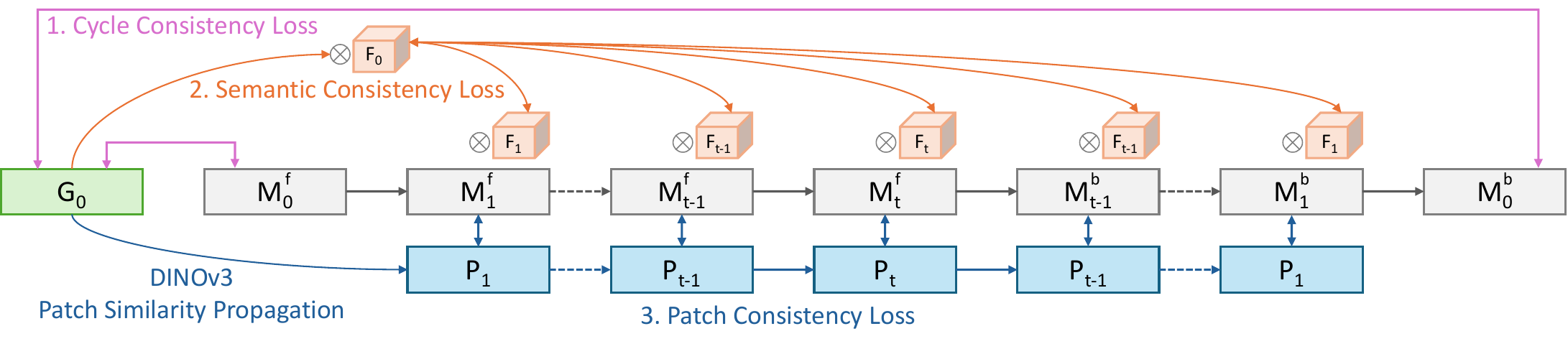}
    \vspace{-0.3cm}
    \caption{Illustration of our label-efficient training strategy. It enables video learning using only the ground-truth mask of the first frame. It consists of (1) a video-level cycle consistency loss, (2) an object-level semantic consistency loss, and (3) a patch-level consistency loss.}
    \label{fig:csp}
    \vspace{-0.3cm}
\end{figure*}

SAM 2~\cite{sam2} requires the large-scale video data with per-frame mask annotation during training. This reliance poses significant challenges for robotic scenarios, where annotating diverse robotic embodiments across dynamic environments is labor-intensive. To alleviate this issue, we propose a label-efficient training strategy that supervises the model using only the ground-truth mask from the first frame of each video. As illustrated in Figure~\ref{fig:csp}, we introduce three consistency losses at different levels of granularity: a video-level cycle consistency loss, an object-level semantic consistency loss, and a patch-level consistency loss.

\textbf{Cycle Consistency Loss.}
A video segmentation model should possess the ability to propagate object masks across frames to ensure temporal consistency. To learn this capability under first-frame-only supervision, we adopt a cycle training strategy. Given a video clip, the model first predicts masks forward from frame $0$ to frame $t$, and then propagates backward from frame $t$ to frame $0$. Since both the initial and final frames correspond to the same time step, their predicted masks, $M_0^{f}$ and $M_0^{b}$, can be supervised by the only available ground-truth mask $G_0$:
\begin{equation}
\mathcal{L}_{\text{cyc}} = \mathcal{D}(M_0^{f}, G_0) + \mathcal{D}(M_0^{b}, G_0),
\end{equation}
where $\mathcal{D}(\cdot, \cdot)$ is a linear combination of focal loss and dice loss. This cycle loss encourages the model to learn cross-frame mask propagation using supervision from the first frame only.

\textbf{Semantic Consistency Loss.}
The cycle consistency loss constrains the segmentation at the first/last frame but provides no supervision for intermediate frames. To complement it, we introduce a semantic consistency loss that enforces object-level coherence across frames and discourages the model from trivially preserving static masks between the start and end frames in dynamic scenes. For each intermediate prediction in $\{M_1^f, \dots, M_t^f, \dots, M_1^b\}$, we compute the object semantic embedding $\mathbf{f}_x$ by spatially averaging the image features $F_x$ within the predicted mask $M_x$, and align it with the first-frame object semantics $\mathbf{f}_0$:
\begin{equation}
\mathcal{L}_{\text{sem}} = 1 - \frac{1}{|\mathcal{T}|} \sum_{x \in \mathcal{T}} 
\left( \frac{\mathbf{f}_x \cdot \mathbf{f}_0}{\|\mathbf{f}_x\| \, \|\mathbf{f}_0\|} \right),
\end{equation}
where $\mathcal{T}$ denotes the set of all intermediate frames. This loss encourages the predicted masks to remain semantically aligned with the target object throughout the video.

\begin{figure}[t]
    \centering
    \includegraphics[width=1\linewidth]{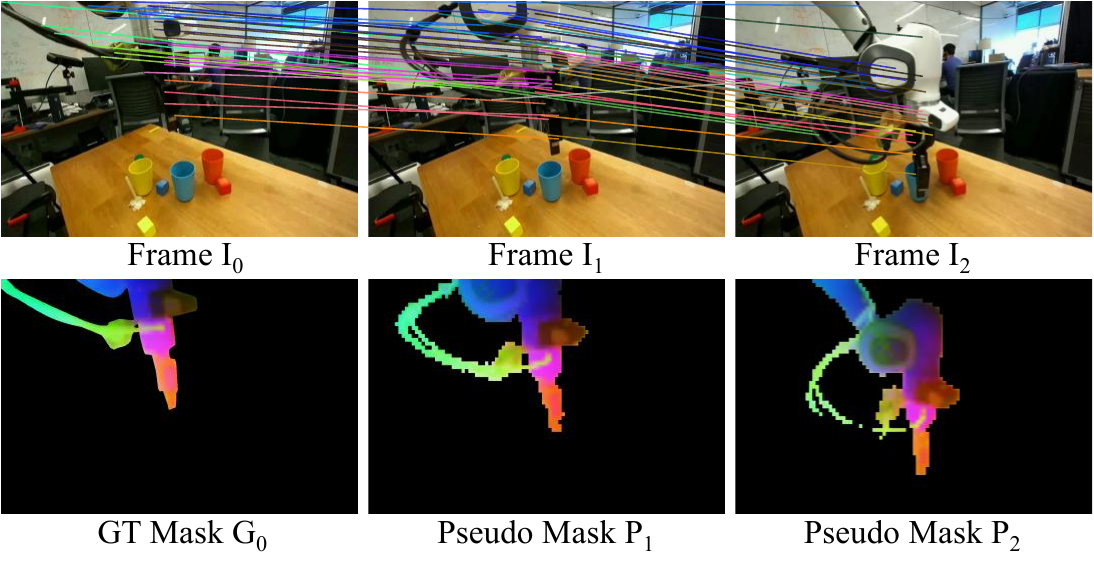}
    \vspace{-0.75cm}
    \caption{Visualization of pseudo-label generation using DINOv3~\cite{simeoni2025dinov3} patch similarity. The top row shows sparse correspondences across frames, while the bottom row visualizes the ground-truth mask $G_0$ and the propagated pseudo masks $P_1$ and $P_2$, colored by PCA-projected features.}
    \label{fig:dinov3-pseudo}
    \vspace{-0.3cm}
\end{figure}

\textbf{Patch Consistency Loss.}
To further enhance spatial alignment and structural fidelity, we introduce a patch-level consistency loss. It supervises the predicted masks $M_x$ using pseudo labels $P_x$ propagated from the first-frame ground truth via DINOv3~\cite{simeoni2025dinov3} patch similarity. The loss is computed as:
\begin{equation}
\mathcal{L}_{\text{patch}} = \frac{1}{|\mathcal{T}|} \sum_{x \in \mathcal{T}} \mathcal{D}'(M_x^{\downarrow 16}, P_x),
\end{equation}
where $^{\downarrow 16}$ denotes downsampling the mask by a factor of 16 to match the patch granularity of DINOv3 features, and $\mathcal{D}'(\cdot, \cdot)$ represents an IoU loss. Figure~\ref{fig:dinov3-pseudo} illustrates an example where we leverage DINOv3 \cite{simeoni2025dinov3} to generate usable pseudo masks through patch-level similarity.
Finally, the mask loss is defined as a weighted sum of the three hierarchical losses:
\begin{equation}
\mathcal{L}_{\text{mask}} 
= w_{\text{cyc}} \mathcal{L}_{\text{cyc}} 
+ w_{\text{sem}} \mathcal{L}_{\text{sem}} 
+ w_{\text{patch}} \mathcal{L}_{\text{patch}},
\end{equation}
where $w_{\text{cyc}}$, $w_{\text{sem}}$, and $w_{\text{patch}}$ are balancing weights.

%% file: sec/5_experiments.tex
\section{Experiments}
\label{experiments}
\subsection{Experimental Setup}
\noindent\textbf{Training Configurations.}
Our RobotSeg is implemented in PyTorch~\cite{paszke2019pytorch} and trained jointly on 3,532 images from RoboEngine-Train~\cite{yuan2025roboengine} and 2,707 videos (131,504 frames) from our VRS-Train dataset. The training is conducted for 25 epochs in 15 hours on 8 NVIDIA RTX~A5000 GPUs with 24\,GB of memory per GPU. We adopt the AdamW optimizer~\cite{loshchilov2017decoupled} with an initial learning rate of $3{\times}10^{-4}$ for the image encoder and $6{\times}10^{-5}$ for the remaining components. All learning rates are scheduled using a cosine decay policy.

\noindent\textbf{Evaluation Settings.}
To comprehensively assess the capabilities of our RobotSeg, we perform comparisons on both the existing image dataset RoboEngine-Test~\cite{yuan2025roboengine} and our video dataset VRS-Test, under five evaluation settings: automatic (\textit{AU}), 1-click (\textit{1C}), 3-click (\textit{3C}), bounding-box (\textit{BB}), and online-interactive (\textit{OI}). Specifically, \textit{AU} denotes automatic segmentation without any click or bounding-box prompts; \textit{1C}, \textit{3C}, and \textit{BB} involve providing a single click, three clicks, or a bounding box on the first frame as prompts. For the \textit{OI} setting, we follow the online interaction protocol of SAM 2 \cite{sam2}, where segmentation starts from a 3-click prompt on the first frame and propagates forward. During propagation, whenever the IoU of a predicted mask drops below a predefined threshold (0.9), additional corrective clicks (up to 3) are introduced, with at most $N{=}3$ such interactive rounds allowed. For all evaluations, we report the standard $J\&F$ metric~\cite{pont20172017}, which combines region similarity (Jaccard index) and boundary accuracy (F-measure).

\begin{figure*}[t]
    \centering
    \includegraphics[width=1\linewidth]{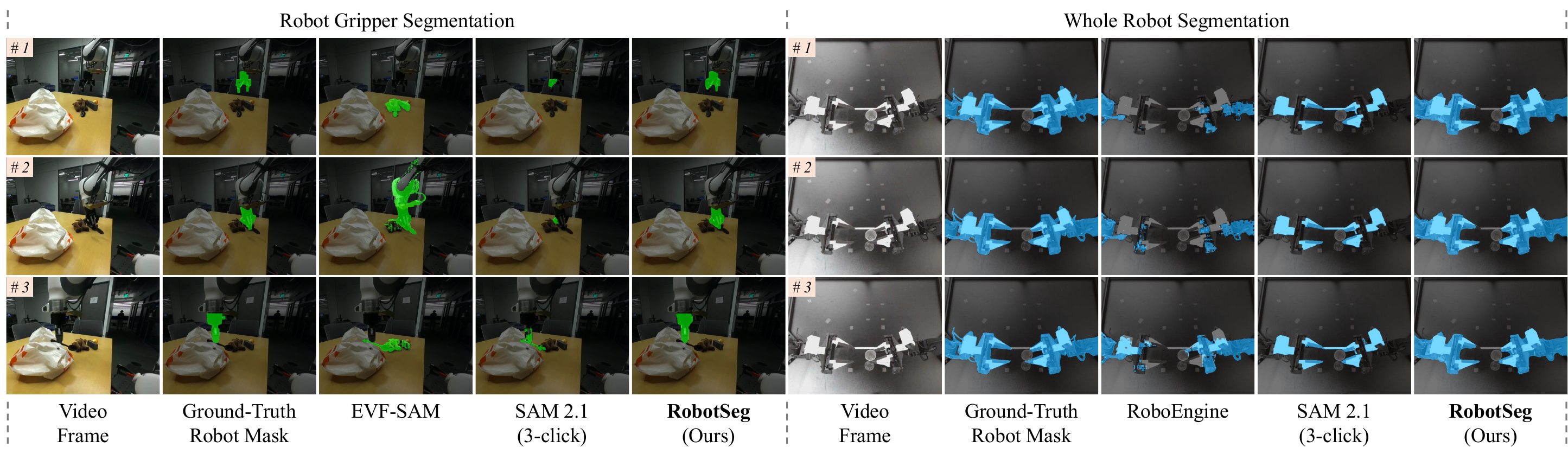}
    \vspace{-0.75cm}
    \caption{Qualitative comparisons on the robot gripper segmentation (left) and the whole robot segmentation (right).}
    \label{fig:visual}
    \vspace{-0.1cm}
\end{figure*}

\begin{table*}[tbp]
    \caption{Comparisons of the video robot segmentation under five settings (\ie, automatic \textit{AU}, 1-click \textit{1C}, 3-click \textit{3C}, bounding-box \textit{BB}, and online-interactive \textit{OI}) on the VRS dataset. ``–'' denotes that the method does not support this setting.}
    \vspace{-5pt}
    \label{tab:vrs}
    \centering
    \footnotesize
    \renewcommand{\arraystretch}{1.05}
    \setlength{\tabcolsep}{4.6pt}
    \begin{tabular}{l|r|c|c|c|c|c|c|c|c|c|c|c|c|c|c|c}
        \hline
        \hline
        \multirow{2}{*}{Methods} & \multirow{2}{*}{\makecell{Para.\\(M)}} & \multicolumn{5}{c|}{Robot Arm} & \multicolumn{5}{c|}{Robot Gripper} & \multicolumn{5}{c}{Whole Robot} \\
        \cline{3-17}
        & & AU & 1C & 3C & BB & OI & AU & 1C & 3C & BB & OI & AU & 1C & 3C & BB & OI \\ 
        \hline
        \hline
        RoVi-Aug \cite{chen2025rovi} & 638.5 & - & - & - & - & - & - & - & - & - & - & 38.9 & - & - & - & - \\
        \hline
        RoboEngine \cite{yuan2025roboengine} (Original) & 898.4 & - & - & - & - & - & - & - & - & - & - & 74.1 & - & - & - & - \\
        \hline
        RoboEngine \cite{yuan2025roboengine} (Finetuned) & 898.4 & - & - & - & - & - & - & - & - & - & - & 80.2 & - & - & - & - \\
        \hline
        \hline
        CLIPSeg \cite{luddecke2022image} & 150.8 & 23.7 & - & - & - & - & 6.7 & - & - & - & - & 26.6 & - & - & - & - \\
        \hline
        LISA \cite{lai2024lisa} & 13992.9 & 42.1 & - & - & - & - & 21.2 & - & - & - & - & 55.3 & - & - & - & - \\
        \hline
        EVF-SAM \cite{zhang2024evf} & 898.4 & 44.7 & - & - & - & - & 23.8 & - & - & - & - & 63.9 & - & - & - & - \\
        \hline
        \hline
        SAM 2.1 \cite{sam2} (Original) & 39.0 & - & 28.9 & 57.1 & 50.2 & 61.8 & - & 47.7 & 65.8 & 67.9 & 69.1 & - & 38.2 & 69.0 & 60.4 & 73.6 \\ \hline
        SAM 2.1 \cite{sam2} (Finetuned) & 39.0 & - & 66.2 & 72.1 & 71.5 & 74.5 & - & 64.8 & 68.4 & 67.8 & 70.4 & - & 73.6 & 82.1 & 82.5 & 85.1 \\ \hline
        \hline
        \textbf{RobotSeg} (Ours) & 41.3 & \textbf{75.6} & \textbf{75.5} & \textbf{76.6} & \textbf{76.4} & \textbf{77.6} & \textbf{76.0} & \textbf{76.3} & \textbf{76.8} & \textbf{76.7} & \textbf{76.9} & \textbf{85.1} & \textbf{85.1} & \textbf{86.3} & \textbf{85.8} & \textbf{86.7} \\
        \hline
        \hline
    \end{tabular}
    \vspace{-3mm}
\end{table*}

\begin{table}[tbp]
    \caption{Comparisons of the image robot segmentation under four settings (\ie, automatic \textit{AU}, 1-click \textit{1C}, 3-click \textit{3C}, and bounding-box \textit{BB}) on the RoboEngine dataset \cite{yuan2025roboengine}. ``–'' denotes that the method does not support this setting.}
    \vspace{-4pt}
    \label{tab:roboengine}
    \centering
    \footnotesize
    \renewcommand{\arraystretch}{1}
    \setlength{\tabcolsep}{4.8pt}
    \begin{tabular}{l|r|c|c|c|c}
        \hline
        \hline
        \multirow{2}{*}{Methods} & \multirow{2}{*}{\makecell{Para.\\(M)}} & \multicolumn{4}{c}{Whole Robot} \\
        \cline{3-6}
        & & AU & 1C & 3C & BB \\
        \hline
        \hline
        RoVi-Aug \cite{chen2025rovi} & 638.5 & 36.1 & - & - & -  \\
        \hline
        RoboEngine \cite{yuan2025roboengine} (Original) & 898.4 & 85.9 & - & - & - \\
        \hline
        RoboEngine \cite{yuan2025roboengine} (Finetuned) & 898.4 & 86.6 & - & - & - \\
        \hline
        \hline
        CLIPSeg \cite{luddecke2022image} & 150.8 & 22.5 & - & - & - \\
        \hline
        LISA \cite{lai2024lisa} & 13992.9 & 54.4 & - & - & - \\
        \hline
        EVF-SAM \cite{zhang2024evf} & 898.4 & 69.2 & - & - & -  \\
        \hline
        \hline
        SAM 2.1 \cite{sam2} (Original) & 39.0 & - & 55.9 & 88.4 & 79.6  \\
        \hline
        SAM 2.1 \cite{sam2} (Finetuned) & 39.0 & - & 78.0 & 90.2 & 86.0 \\
        \hline
        \hline
        \textbf{RobotSeg} (Ours) & 41.3 & \textbf{87.9} & \textbf{88.8} & \textbf{93.5} & \textbf{89.4} \\
        \hline
        \hline
    \end{tabular}
    \vspace{-4mm}
\end{table}

\subsection{Comparison to Existing Methods}
From Table~\ref{tab:vrs},~\ref{tab:roboengine} and Figure~\ref{fig:visual}, we draw the following conclusions:
(i) our RobotSeg outperforms existing robot segmentation methods RoVi-Aug~\cite{chen2025rovi} and RoboEngine~\cite{yuan2025roboengine} by a significant margin ({$\geq$}4.9 $J\&F$) in the automatic setting, and uniquely provides finer-grained masks for the robot arm and gripper, enabling potential applications such as part-aware data augmentation. Moreover, RobotSeg supports prompt-based segmentation, which further improves accuracy over the automatic mode and is particularly beneficial in scenarios where user-provided prompts are available to achieve optimal performance;
(ii) compared with language-conditioned segmentation models CLIPSeg~\cite{luddecke2022image}, LISA~\cite{lai2024lisa} and EVF-SAM~\cite{zhang2024evf}, RobotSeg yields substantially stronger automatic segmentation;
(iii) compared with the state-of-the-art promptable video foundation model SAM~2.1~\cite{sam2}, RobotSeg achieves markedly higher accuracy under all prompting settings due to its robot-specific design;
(iv) same conclusions hold on the image dataset RoboEngine~\cite{yuan2025roboengine}, confirming the strong capability of RobotSeg for both video and image robot segmentation;
(v) qualitative comparisons in Figure~\ref{fig:visual} show that RobotSeg produces more accurate and structurally consistent masks;
and (vi) RobotSeg contains only 41.3M parameters, which is significantly smaller than existing robot segmentation models ({$\geq$}638.5M) and language-conditioned models ({$\geq$}150.8M), making it suitable for real-world deployment where efficiency and compactness are essential.

\subsection{Ablation Study}
\noindent\textbf{Impact of Robot-Specific Fine-Tuning.}
Comparing Table~\ref{tab:ablation} (a) and (b), we observe a substantial improvement from 38.2 to 73.6 $J\&F$ after fine-tuning the model on our robot dataset, showing the importance of domain adaptation for robot segmentation and 
highlighting the value of our dataset in enabling such adaptation.

\noindent\textbf{Influence of Label-Efficient Training (LET).}
Variants (c)–(e) progressively add the cycle, semantic, and patch-level consistency losses. Each component brings steady gains, with the full LET (e) achieving 77.4 $J\&F$, validating the effectiveness of our hierarchical supervision strategy using only the first-frame ground truth.

\noindent\textbf{Effectiveness of RPG.}
Introducing class tokens (f) enables autonomous robot segmentation by providing class-level semantic priors. Incorporating object tokens (g) further improves performance by leveraging memory-derived object cues for more precise and temporally consistent guidance.

\noindent\textbf{Effectiveness of SEMA.}
Beyond the original memory association in SAM 2 \cite{sam2}, our added structure enhancement branch brings clear gains. The multi-scale perception (h) and the memory-guided structure modulation (i) together boosts performance to 85.1, demonstrating the value of structure enhancement for articulated robot segmentation.

\begin{table}[t]
\caption{Ablation studies. “\cmark” indicates the component is enabled. “--” denotes unavailable results.}\vspace{-2mm}
\label{tab:ablation}
\centering
\footnotesize
\renewcommand{\arraystretch}{.95}
\setlength{\tabcolsep}{3.2pt}
\begin{tabular}{c|c|c|c|c|c|c|c|c|c|c}
\hline
\hline
\multirow{2}{*}{Variants} & \multirow{2}{*}{FT} &
\multicolumn{3}{c|}{LET} &
\multicolumn{2}{c|}{RPG} &
\multicolumn{2}{c|}{SEMA} &
\multirow{2}{*}{AU} & \multirow{2}{*}{1C} \\
\cline{3-5} \cline{6-7} \cline{8-9}
 & & Cyc. & Sem. & Pat. & Cla. & Obj. & MS & MG &  &  \\ 
\hline
(\textit{a}) &  &  &  &  &  &  &  &  & -- & 38.2 \\
(\textit{b}) & \cmark &  &  &  &  &  &  &  & -- & 73.6 \\
\hline
\hline
(\textit{c}) &  & \cmark &  &  &  &  &  &  & -- & 74.7 \\
(\textit{d}) &  & \cmark & \cmark &  &  &  &  &  & -- & 76.3 \\
(\textit{e}) &  & \cmark & \cmark & \cmark &  &  &  &  & -- & 77.4 \\
\hline
\hline
(\textit{f}) &  & \cmark & \cmark & \cmark & \cmark &  &  &  & 79.8 & 79.9 \\
(\textit{g}) &  & \cmark & \cmark & \cmark & \cmark & \cmark &  &  & 83.1 & 83.3 \\
\hline
\hline
(\textit{h}) &  & \cmark & \cmark & \cmark & \cmark & \cmark & \cmark &  & 83.7 & 83.8 \\
(\textit{i}) &  & \cmark & \cmark & \cmark & \cmark & \cmark & \cmark & \cmark & 85.1 & 85.1 \\
\hline
\hline
\end{tabular}
\vspace{-4mm}
\end{table}

%% file: sec/6_conclusion.tex
\section{Conclusion}
\label{conclusion}
In this work, we presented RobotSeg, the first foundation model for robot segmentation that supports both images and videos. Built upon the versatile SAM 2 model, RobotSeg addresses the unique challenges of robotic perception by introducing three key innovations: a structure-enhanced memory associator for temporally consistent and structure-aware segmentation, a robot prompt generator for autonomous mask prediction, and a label-efficient training strategy that enables supervision with only the first-frame mask annotation. We also constructed the VRS dataset, the first video robot segmentation benchmark spanning diverse robot embodiments and environments. Extensive experiments on both images and videos demonstrate that RobotSeg achieves state-of-the-art performance, establishing a strong foundation for future advances in robot perception and enabling more reliable robot-centric downstream tasks.

\noindent\textbf{Acknowledgments.} This project is supported by the National Research Foundation, Singapore under its NRFF award NRF-NRFF13-2021-0008.

%% file: sec/7_supp.tex
\clearpage
\maketitlesupplementary

\section{Overview}
\label{sec:overview}
In this supplementary material, we provide additional experiments, analyses, and implementation details to complement the main paper. We first present an application example of RobotSeg for robot-centric data augmentation in Section \ref{sec:aug}. We then show more dataset examples from our VRS benchmark in Section \ref{sec:dataset_examples}. Next, we report comprehensive category-wise evaluations across robot embodiments in Section \ref{sec:class_comparison}, highlighting the robustness of RobotSeg across diverse robot embodiments. We include more visual comparisons with state-of-the-art methods in Section \ref{sec:visual_comparison} and provide computational efficiency analyses in Section \ref{sec:flops}. We further provide an additional comparison with SAM\,3 \cite{sam3} for completeness in Section \ref{sec:sam3}. Finally, detailed architecture descriptions of key modules, including the memory encoder, the structure perceiver, and the mask decoder, are provided in Section \ref{sec:technical_details}, followed by discussions on limitations and future directions in Section \ref{sec:limitations}.

\section{An Application Example of RobotSeg}
\label{sec:aug}
Accurate robot segmentation is a prerequisite for reliable robot perception and tracking, and it further enables the creation of high-quality training data for robot learning. A practical example is robot-centric data augmentation, where robot masks are used to composite the robot into diverse scenes to improve the robustness of downstream policies. However, this process is highly sensitive to mask quality: segmentation errors such as missing parts, drifted boundaries, or broken structures directly translate into unrealistic composites that can harm the learning signal.

\begin{figure*}[t]
    \centering
    \includegraphics[width=1\linewidth]{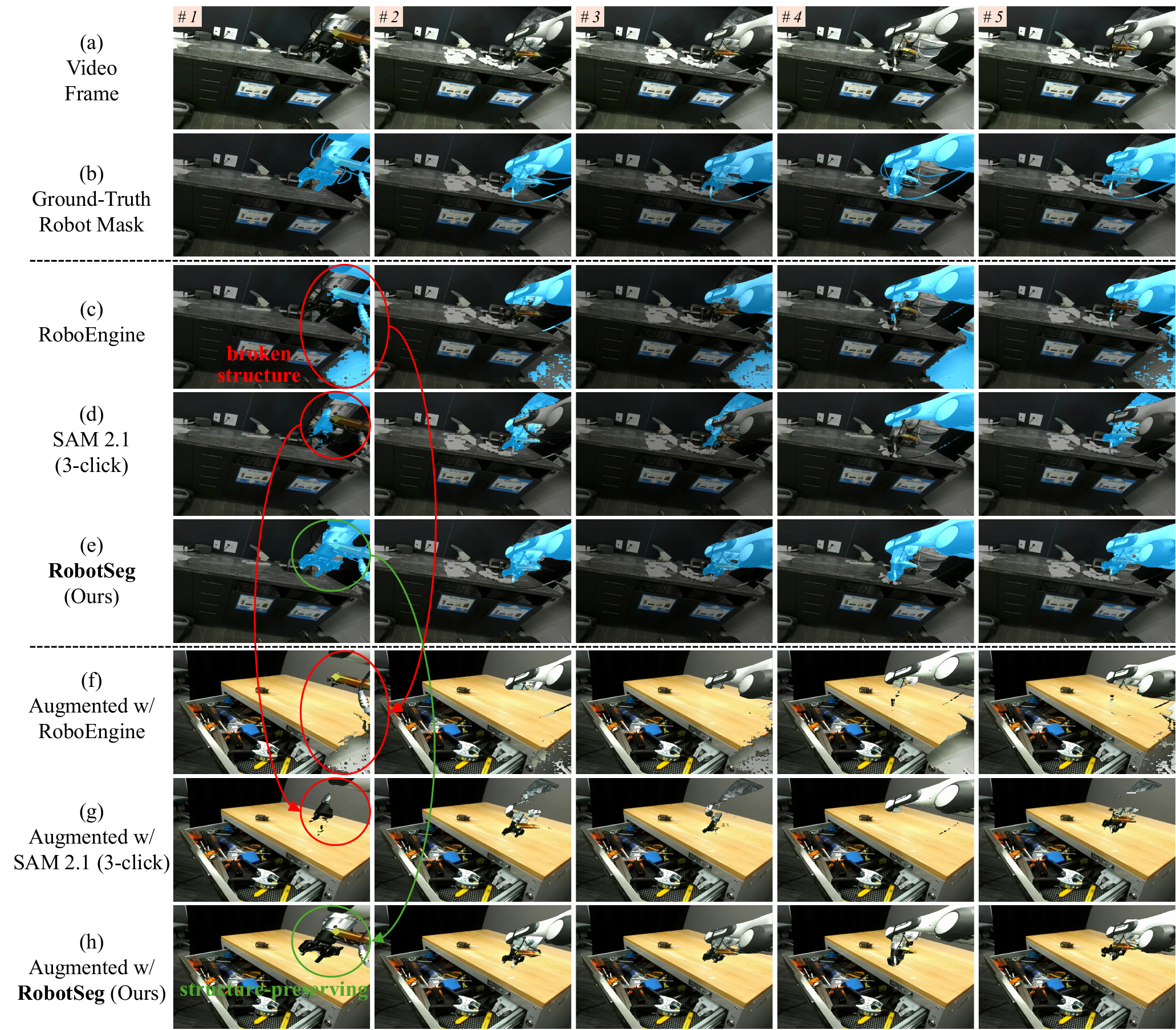}
    \caption{Robot segmentations obtained with the image robot segmentation method RoboEngine \cite{yuan2025roboengine} (c) and the promptable video segmentation method SAM 2.1 (with 3 manual clicks to initialize the segmentation) \cite{sam2} (d), compared to our RobotSeg model (e). When used for robot data augmentation, inaccurate masks from RoboEngine and SAM 2.1 lead to broken or unrealistic robot composites (f-g), whereas our RobotSeg enables clean and structurally accurate augmentation (h) by precisely preserving the robot regions.}
    \label{fig:aug}
\end{figure*}

Figure \ref{fig:aug} illustrates this effect on video frames. RoboEngine~\cite{yuan2025roboengine}, which operates on individual images, often produces fragmented or structurally damaged masks (Figure~\ref{fig:aug}c), while SAM~2.1~\cite{sam2} requires manual clicks and still struggles to maintain spatial and temporal consistency across frames (Figure~\ref{fig:aug}d). These inaccuracies propagate to the augmented images, leading to visually implausible robot placements with missing limbs or distorted geometry (Figure~\ref{fig:aug}f-g).

In contrast, RobotSeg generates clean, complete, and temporally stable masks (Figure~\ref{fig:aug}e), enabling high-fidelity robot compositing (Figure~\ref{fig:aug}h). The preserved geometry and consistent boundaries result in realistic augmented images that maintain the structural integrity of the robot. This example highlights the practical value of precise robot segmentation in creating large-scale, diverse, and reliable training data for robot learning systems.

\section{More VRS Dataset Examples}
\label{sec:dataset_examples}
\begin{figure*}[t]
    \centering
    \vspace{6mm}
    \includegraphics[width=1\linewidth]{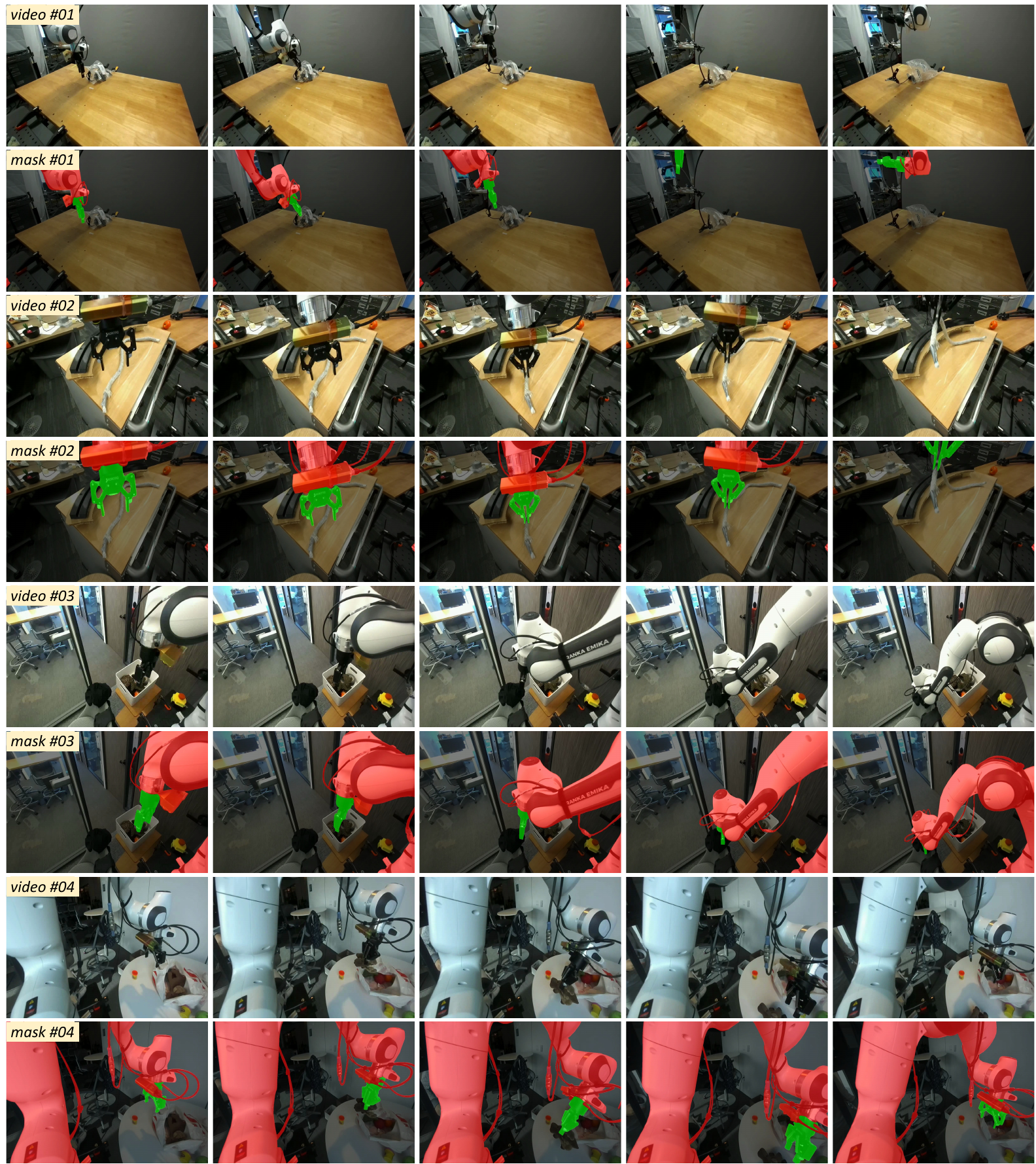}
    \caption{Examples from our video robot segmentation (VRS) dataset. Each example shows the RGB sequence (top) and robot annotation masks (bottom), where the robot arm is highlighted in red and the gripper in green.}
    \label{fig:data1}
\end{figure*}

\begin{figure*}[t]
    \centering
    \vspace{6mm}
    \includegraphics[width=1\linewidth]{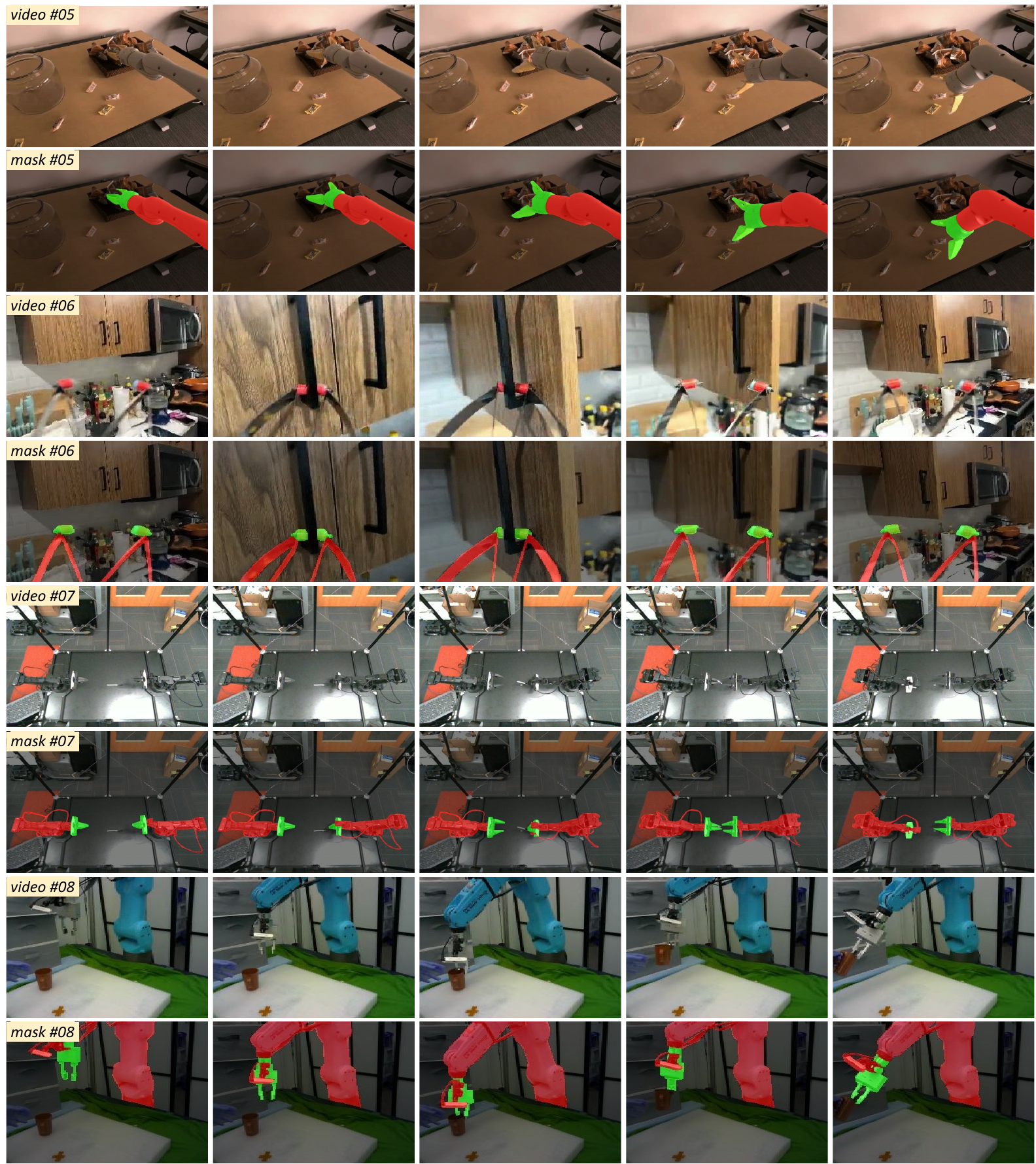}
    \caption{Examples from our video robot segmentation (VRS) dataset. Each example shows the RGB sequence (top) and robot annotation masks (bottom), where the robot arm is highlighted in red and the gripper in green.}
    \label{fig:data2}
\end{figure*}

To complement the dataset overview in the main paper, we present additional examples from our video robot segmentation (VRS) dataset in Figure~\ref{fig:data1} and~\ref{fig:data2}. These examples further illustrate the diversity of robot embodiments, manipulation behaviors, and scene contexts captured in VRS. Each example shows the RGB video frames (top) and their corresponding annotation masks (bottom), where the robot arm and gripper are labeled separately following the hierarchical labeling protocol.

Across these samples, VRS demonstrates substantial variation in motion patterns, viewpoints, backgrounds, object interactions, and lighting conditions. Such diversity is essential for training and evaluating models that aim to achieve robust, temporally consistent robot segmentation in realistic and dynamic environments. By providing continuous video sequences with fine-grained arm and gripper masks, VRS enables research on temporal modeling, mask propagation, and structure-aware segmentation beyond what is possible with the image-only dataset RoboEngine \cite{yuan2025roboengine}.

\vspace{-1pt}
\section{Category-Wise Analysis Across Diverse Robot Embodiments}
\label{sec:class_comparison}
To further evaluate the robustness of different models under diverse robot embodiments, we conduct a category-wise analysis on the VRS dataset. While the main manuscript reports overall results for whole-robot, robot-arm, and robot-gripper segmentation, such aggregated metrics may conceal substantial variations across robots with distinct structures, kinematics, and visual characteristics. Therefore, we additionally present per-category results for all 10 robot embodiments, separately for whole robot (Table~\ref{tab:vrs_robot}), robot arm (Table~\ref{tab:vrs_arm}), and robot gripper (Table~\ref{tab:vrs_gripper}).

\begin{table*}[tbp]
    \caption{Comparison of ``whole robot'' segmentation across the 10 robot categories in the VRS dataset. The superscript ${(1c)}$ denotes that SAM~2.1 \cite{sam2} uses a manual click for initialization.}
    \vspace{-5pt}
    \label{tab:vrs_robot}
    \centering
    \footnotesize
    \renewcommand{\arraystretch}{1.4}
    \setlength{\tabcolsep}{2.8pt}
    \begin{tabular}{l|r|c|c|c|c|c|c|c|c|c|c|c}
    \hline
    \hline
        Methods &
        \makecell{Para.\\(M)} &
        \makecell{\scriptsize{\#01}\\\scriptsize{Franka}} &
        \makecell{\scriptsize{\#02}\\\scriptsize{Fanuc Mate}} &
        \makecell{\scriptsize{\#03}\\\scriptsize{UR5}} &
        \makecell{\scriptsize{\#04}\\\scriptsize{Kuka iiwa}} &
        \makecell{\scriptsize{\#05}\\\scriptsize{Google Robot}} &
        \makecell{\scriptsize{\#06}\\\scriptsize{MobileALOHA}} &
        \makecell{\scriptsize{\#07}\\\scriptsize{xArm}} &
        \makecell{\scriptsize{\#08}\\\scriptsize{WindowX}} &
        \makecell{\scriptsize{\#09}\\\scriptsize{Sawyer}} &
        \makecell{\scriptsize{\#10}\\\scriptsize{Hello Stretch}} &
        \scriptsize{\textbf{Overall}} \\
        \hline
        \hline
        RoVi-Aug \cite{chen2025rovi} & 638.5 & 29.8 & 69.9 & 51.1 & 72.2 & 34.6 & 41.6 & 81.1 & 49.1 & 41.7 & 19.1 & 38.9 \\
        \hline
        RoboEngine \cite{yuan2025roboengine} & 898.4 & 75.8 & \textbf{88.5} & 78.2 & 87.8 & 89.8 & 73.4 & 94.0 & \textbf{91.4} & 83.2 & 6.4 & 74.1 \\
        \hline
        \hline
        CLIPSeg \cite{luddecke2022image} & 150.8 & 20.6 & 38.6 & 32.6 & 59.5 & 33.1 & 29.1 & 52.3 & 29.5 & 35.4 & 15.0 & 26.6 \\
        \hline
        LISA \cite{lai2024lisa} & 13992.9 & 47.1 & 48.6 & 80.0 & 80.5 & 78.3 & 51.8 & 88.6 & 74.4 & 55.4 & 35.2 & 55.3 \\
        \hline
        EVF-SAM \cite{zhang2024evf} & 898.4 & 56.7 & 56.4 & 81.0 & 68.7 & 82.6 & 61.2 & 93.0 & 81.3 & 58.9 & 52.0 & 63.9 \\
        \hline
        VideoLISA \cite{bai2024one} & 4788.3 & 48.9 & 58.8 & 67.5 & 75.3 & 75.9 & 63.4 & 93.5 & 46.4 & 57.2 & 35.7 & 53.6 \\
        \hline
        \hline
        SAM 2.1$^{(1c)}$ \cite{sam2} (Original) & 39.0 & 25.0 & 35.6 & 78.8 & 8.8 & 64.7 & 21.1 & 30.3 & 88.0 & 22.2 & 14.6 & 38.2 \\
        \hline
        SAM 2.1$^{(1c)}$ \cite{sam2} (Finetuned) & 39.0 & 74.7 & 81.4 & 80.4 & \textbf{91.4} & 78.8 & 45.6 & 91.0 & 85.4 & 87.5 & 68.5 & 73.6 \\
        \hline
        \hline
        SAM 3 \cite{sam3} & 860.1 & 30.5 & 59.9 & 0.0 & 70.8 & 19.0 & 41.5 & 92.7 & 60.2 & 8.4 & 0.0 & 34.7 \\
        \hline
        \hline
        \textbf{RobotSeg} (Ours) & 41.3 & \textbf{82.1} & 85.0 & \textbf{88.0} & 85.3 & \textbf{90.2} & \textbf{87.1} & \textbf{95.6} & 87.5 & \textbf{95.3} & \textbf{78.7} & \textbf{85.1} \\
        \hline
        \hline
    \end{tabular}
    \vspace{-1.5mm}
\end{table*}

\begin{table*}[htbp]
    \caption{Comparison of ``robot arm'' segmentation across the 10 robot categories in the VRS dataset. The superscript ${(1c)}$ denotes that SAM~2.1 \cite{sam2} uses a manual click for initialization.}
    \vspace{-5pt}
    \label{tab:vrs_arm}
    \centering
    \footnotesize
    \renewcommand{\arraystretch}{1.4}
    \setlength{\tabcolsep}{2.8pt}
    \begin{tabular}{l|r|c|c|c|c|c|c|c|c|c|c|c}
    \hline
    \hline
        Methods &
        \makecell{Para.\\(M)} &
        \makecell{\scriptsize{\#01}\\\scriptsize{Franka}} &
        \makecell{\scriptsize{\#02}\\\scriptsize{Fanuc Mate}} &
        \makecell{\scriptsize{\#03}\\\scriptsize{UR5}} &
        \makecell{\scriptsize{\#04}\\\scriptsize{Kuka iiwa}} &
        \makecell{\scriptsize{\#05}\\\scriptsize{Google Robot}} &
        \makecell{\scriptsize{\#06}\\\scriptsize{MobileALOHA}} &
        \makecell{\scriptsize{\#07}\\\scriptsize{xArm}} &
        \makecell{\scriptsize{\#08}\\\scriptsize{WindowX}} &
        \makecell{\scriptsize{\#09}\\\scriptsize{Sawyer}} &
        \makecell{\scriptsize{\#10}\\\scriptsize{Hello Stretch}} &
        \scriptsize{\textbf{Overall}} \\
        \hline
        \hline
        CLIPSeg \cite{luddecke2022image} & 150.8 & 23.3 & 45.4 & 32.8 & 65.7 & 22.7 & 26.2 & 44.3 & 10.7 & 22.1 & 19.7 & 23.7 \\
        \hline
        LISA \cite{lai2024lisa} & 13992.9 & 45.8 & 57.0 & 62.1 & 75.2 & 54.4 & 39.6 & 90.9 & 24.9 & 45.5 & 25.1 & 42.1 \\
        \hline
        EVF-SAM \cite{zhang2024evf} & 898.4 & 51.3 & 56.2 & 66.2 & 76.4 & 56.5 & 38.6 & 91.5 & 27.5 & 41.1 & 29.3 & 44.7 \\
        \hline
        VideoLISA \cite{bai2024one} & 4788.3 & 39.5 & 45.1 & 58.5 & 64.5 & 54.9 & 43.2 & 89.9 & 26.5 & 37.5 & 48.5 & 41.4 \\
        \hline
        \hline
        SAM 2.1$^{(1c)}$ \cite{sam2} (Original) & 39.0 & 26.3 & 41.4 & 52.9 & 9.1 & 58.4 & 16.4 & 31.7 & 30.6 & 26.9 & 18.7 & 28.9 \\
        \hline
        SAM 2.1$^{(1c)}$ \cite{sam2} (Finetuned) & 39.0 & \textbf{79.0} & 79.0 & 81.2 & \textbf{87.1} & 60.5 & 38.7 & 87.2 & 54.5 & 72.4 & 62.5 & 66.2 \\
        \hline
        \hline
        SAM 3 \cite{sam3} & 860.1 & 51.9 & 62.4 & 74.0 & 69.6 & 64.3 & 52.2 & 89.4 & 26.8 & 39.4 & 0.0 & 45.0 \\
        \hline
        \hline
        \textbf{RobotSeg} (Ours) & 41.3 & 76.6 & \textbf{79.8} & \textbf{83.4} & 82.9 & \textbf{85.6} & \textbf{72.2} & \textbf{92.4} & \textbf{61.2} & \textbf{95.0} & \textbf{76.0} & \textbf{75.6} \\
        \hline
        \hline
    \end{tabular}
    \vspace{-1.5mm}
\end{table*}

\begin{table*}[htbp]
    \caption{Comparison of ``robot gripper'' segmentation across the 10 robot categories in the VRS dataset. The superscript ${(1c)}$ denotes that SAM~2.1 \cite{sam2} uses a manual click for initialization.}
    \vspace{-5pt}
    \label{tab:vrs_gripper}
    \centering
    \footnotesize
    \renewcommand{\arraystretch}{1.4}
    \setlength{\tabcolsep}{2.8pt}
    \begin{tabular}{l|r|c|c|c|c|c|c|c|c|c|c|c}
    \hline
    \hline
        Methods &
        \makecell{Para.\\(M)} &
        \makecell{\scriptsize{\#01}\\\scriptsize{Franka}} &
        \makecell{\scriptsize{\#02}\\\scriptsize{Fanuc Mate}} &
        \makecell{\scriptsize{\#03}\\\scriptsize{UR5}} &
        \makecell{\scriptsize{\#04}\\\scriptsize{Kuka iiwa}} &
        \makecell{\scriptsize{\#05}\\\scriptsize{Google Robot}} &
        \makecell{\scriptsize{\#06}\\\scriptsize{MobileALOHA}} &
        \makecell{\scriptsize{\#07}\\\scriptsize{xArm}} &
        \makecell{\scriptsize{\#08}\\\scriptsize{WindowX}} &
        \makecell{\scriptsize{\#09}\\\scriptsize{Sawyer}} &
        \makecell{\scriptsize{\#10}\\\scriptsize{Hello Stretch}} &
        \scriptsize{\textbf{Overall}} \\
        \hline
        \hline
        CLIPSeg \cite{luddecke2022image} & 150.8 & 2.4 & 8.3 & 1.9 & 0.9 & 6.7 & 10.1 & 2.9 & 15.2 & 2.7 & 5.8 & 6.7 \\
        \hline
        LISA \cite{lai2024lisa} & 13992.9 & 9.3 & 7.0 & 26.4 & 6.8 & 35.1 & 29.2 & 2.9 & 43.5 & 33.1 & 12.9 & 21.2 \\
        \hline
        EVF-SAM \cite{zhang2024evf} & 898.4 & 11.4 & 27.0 & 22.6 & 10.1 & 33.8 & 34.0 & 10.9 & 38.8 & 38.6 & 15.0 & 23.8 \\
        \hline
        VideoLISA \cite{bai2024one} & 4788.3 & 4.5 & 2.2 & 19.0 & 1.5 & 34.3 & 29.7 & 4.4 & 29.4 & 27.4 & 8.0 & 15.9 \\
        \hline
        \hline
        SAM 2.1$^{(1c)}$ \cite{sam2} (Original) & 39.0 & 52.0 & 35.6 & 42.1 & 21.8 & 46.7 & 15.3 & \textbf{88.0} & 72.1 & 48.0 & 48.3 & 47.7 \\
        \hline
        SAM 2.1$^{(1c)}$ \cite{sam2} (Finetuned) & 39.0 & 52.8 & \textbf{70.0} & 53.8 & \textbf{66.8} & 59.6 & 70.3 & 18.6 & 79.3 & 76.1 & \textbf{78.6} & 64.8 \\
        \hline
        \hline
        SAM 3 \cite{sam3} & 860.1 & 11.9 & 0.5 & 9.3 & 5.8 & 25.0 & 11.6 & 10.0 & 5.9 & 17.0 & 5.6 & 10.5 \\
        \hline
        \hline
        \textbf{RobotSeg} (Ours) & 41.3 & \textbf{71.6} & 62.4 & \textbf{68.2} & 64.6 & \textbf{80.7} & \textbf{78.7} & 78.8 & \textbf{85.0} & \textbf{87.5} & 77.4 & \textbf{76.0} \\
        \hline
        \hline
    \end{tabular}
    \vspace{-1.5mm}
\end{table*}

Across the three segmentation targets, RobotSeg consistently ranks among the top-performing methods within each robot embodiment. RobotSeg ranks first in 7, 8, and 6 out of 10 robot types for segmenting the whole robot (Table~\ref{tab:vrs_robot}), robot arm (Table~\ref{tab:vrs_arm}), and robot gripper (Table~\ref{tab:vrs_gripper}), respectively. These results indicate that RobotSeg maintains stable performance across diverse robot embodiments with different shapes, sizes, and visual characteristics. In contrast, existing approaches, including the \textbf{\textit{robot-specific methods}} RoVi-Aug~\cite{chen2025rovi} and RoboEngine~\cite{yuan2025roboengine}, the \textbf{\textit{language-conditioned models}} CLIPSeg~\cite{luddecke2022image}, LISA~\cite{lai2024lisa}, EVF-SAM~\cite{zhang2024evf}, and VideoLISA \cite{bai2024one}, the \textbf{\textit{promptable video foundation model}} SAM~2.1~\cite{sam2} (with one manual click prompt to initialize the segmentation), and the \textbf{\textit{concept segmentation model}} SAM 3 \cite{sam3}, show large variation across robot categories. The category-wise analysis indicates that RobotSeg generalizes more reliably across real-world robotic embodiments than existing models.

These results highlight two key insights. First, robot segmentation is highly embodiment-dependent: strong average performance does not guarantee robustness across distinct robot embodiments, as illustrated by SAM~2.1 (finetuned) whose robot-arm accuracy is 66.2 overall but drops to 38.7 on the MobileALOHA robot (Table~\ref{tab:vrs_arm}). Second, RobotSeg’s robot-aware design and label-efficient video training enable generalization across diverse embodiments for whole-robot, arm-level, and gripper-level segmentation. 
This demonstrates that RobotSeg is not only favorable on average but also reliably transferable to a wide spectrum of real-world robotic embodiments.

\section{More Visual Comparison Results}
\label{sec:visual_comparison}
\begin{figure*}[t]
    \centering
    \vspace{6mm}
    \includegraphics[width=1\linewidth]{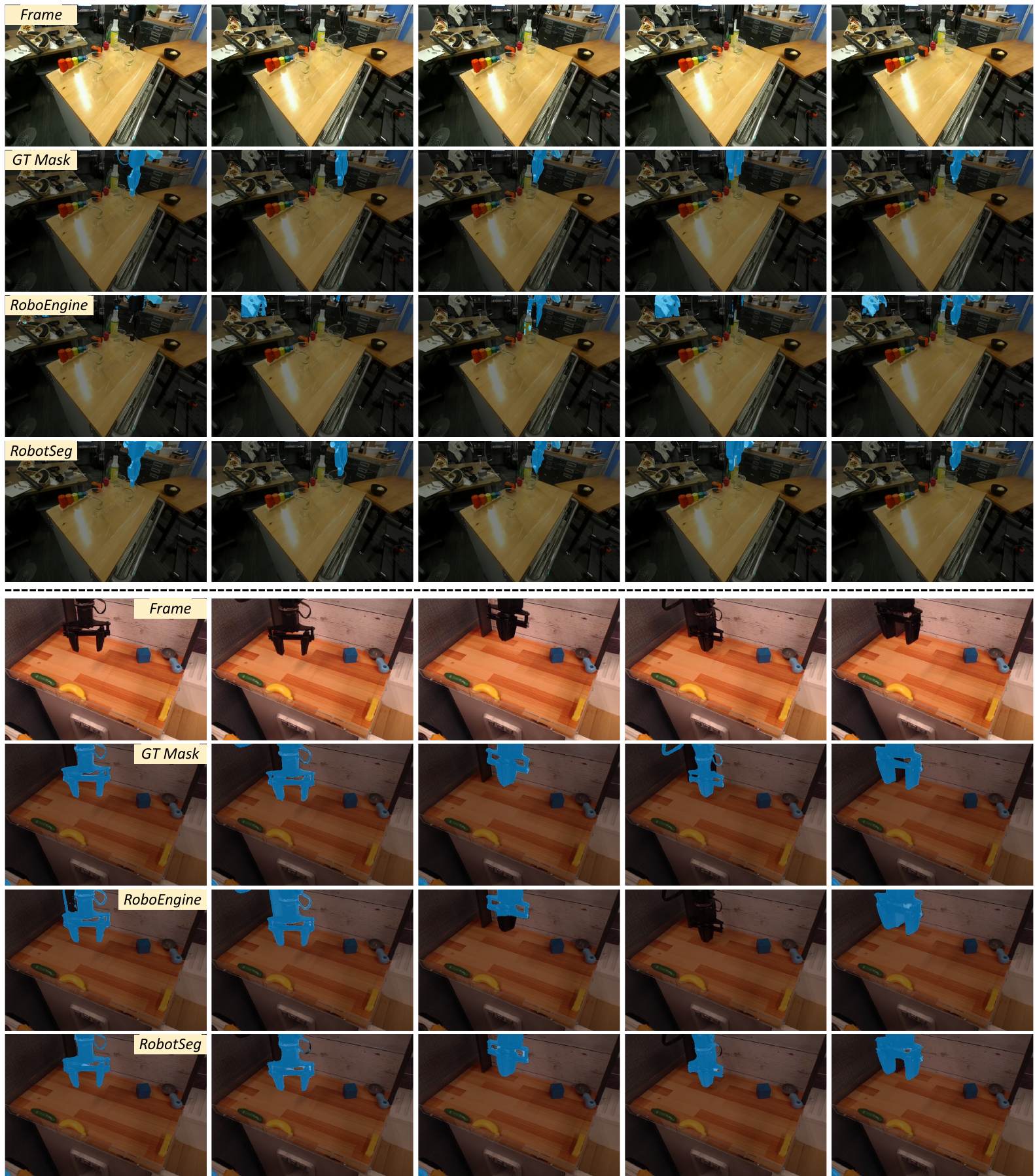}
    \caption{Qualitative comparison of whole-robot segmentation under the automatic setting. RoboEngine \cite{yuan2025roboengine} exhibits clear inaccuracies and temporal inconsistencies, while our RobotSeg produces accurate and stable masks across frames.}
    \label{fig:visual_robot}
\end{figure*}

\begin{figure*}[t]
    \centering
    \vspace{6mm}
    \includegraphics[width=1\linewidth]{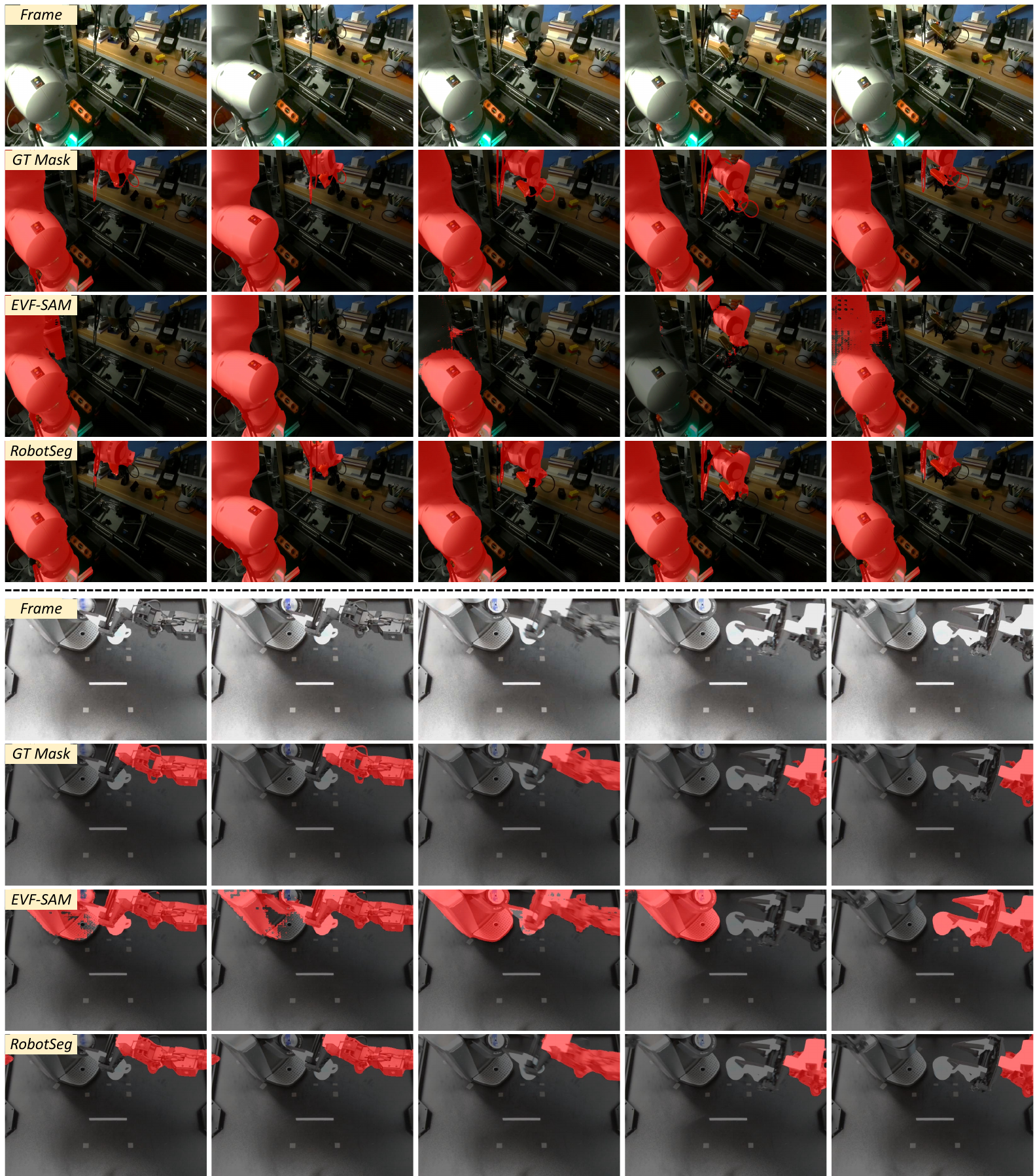}
    \caption{Qualitative comparison of robot arm segmentation under the automatic setting. EVF-SAM \cite{zhang2024evf} struggles to localize the articulated arm and often confuses background objects, whereas our RobotSeg provides accurate, component-specific, and temporally consistent predictions.}
    \label{fig:visual_arm}
\end{figure*}

\begin{figure*}[t]
    \centering
    \vspace{6mm}
    \includegraphics[width=1\linewidth]{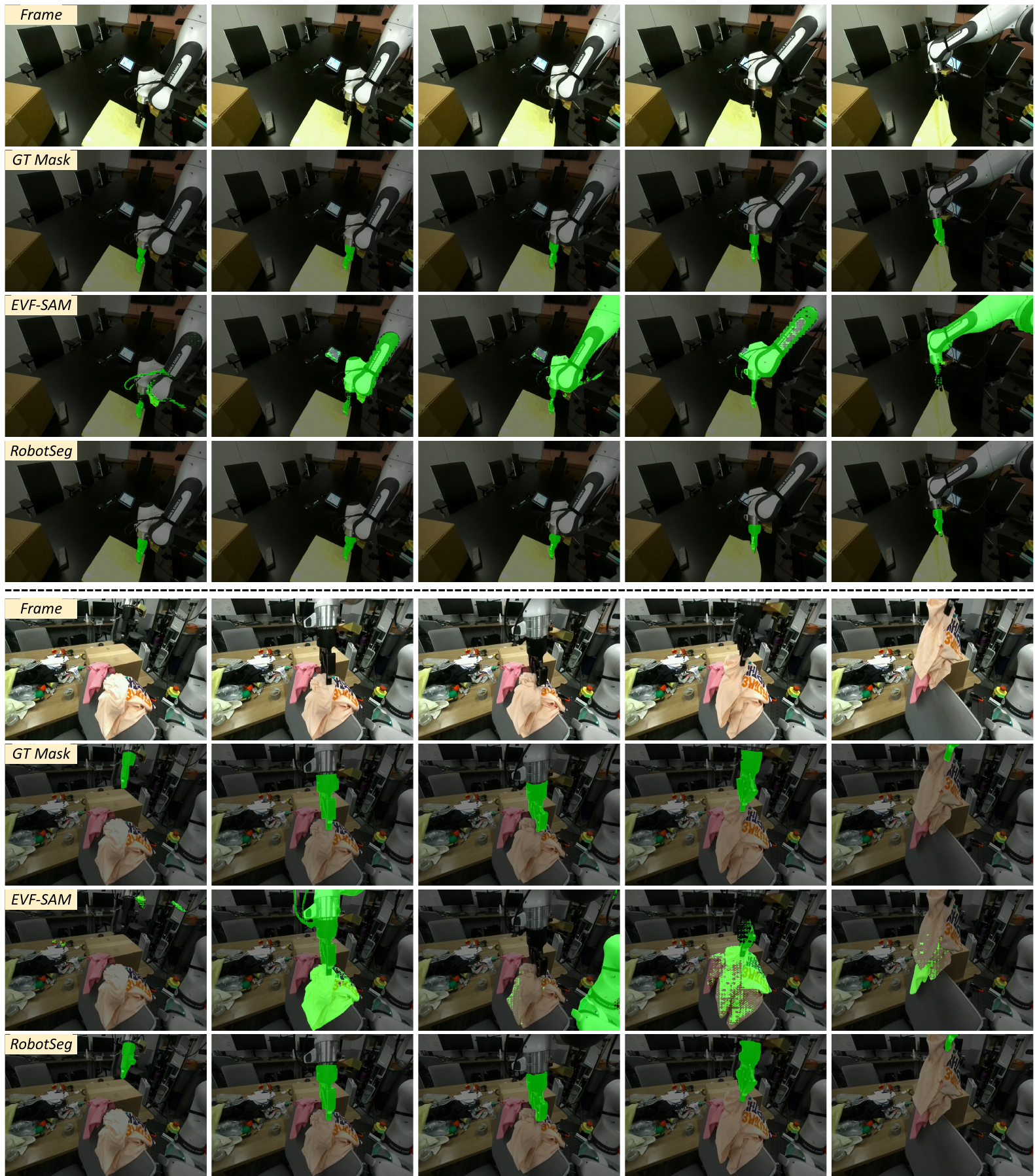}
    \caption{Qualitative comparison of robot gripper segmentation under the automatic setting. EVF-SAM \cite{zhang2024evf} frequently misidentifies the gripper or mistakes background regions for the target, while our RobotSeg consistently segments the correct component with stable temporal behavior.}
    \label{fig:visual_gripper}
\end{figure*}

\begin{figure*}[t]
    \centering
    \vspace{6mm}
    \includegraphics[width=1\linewidth]{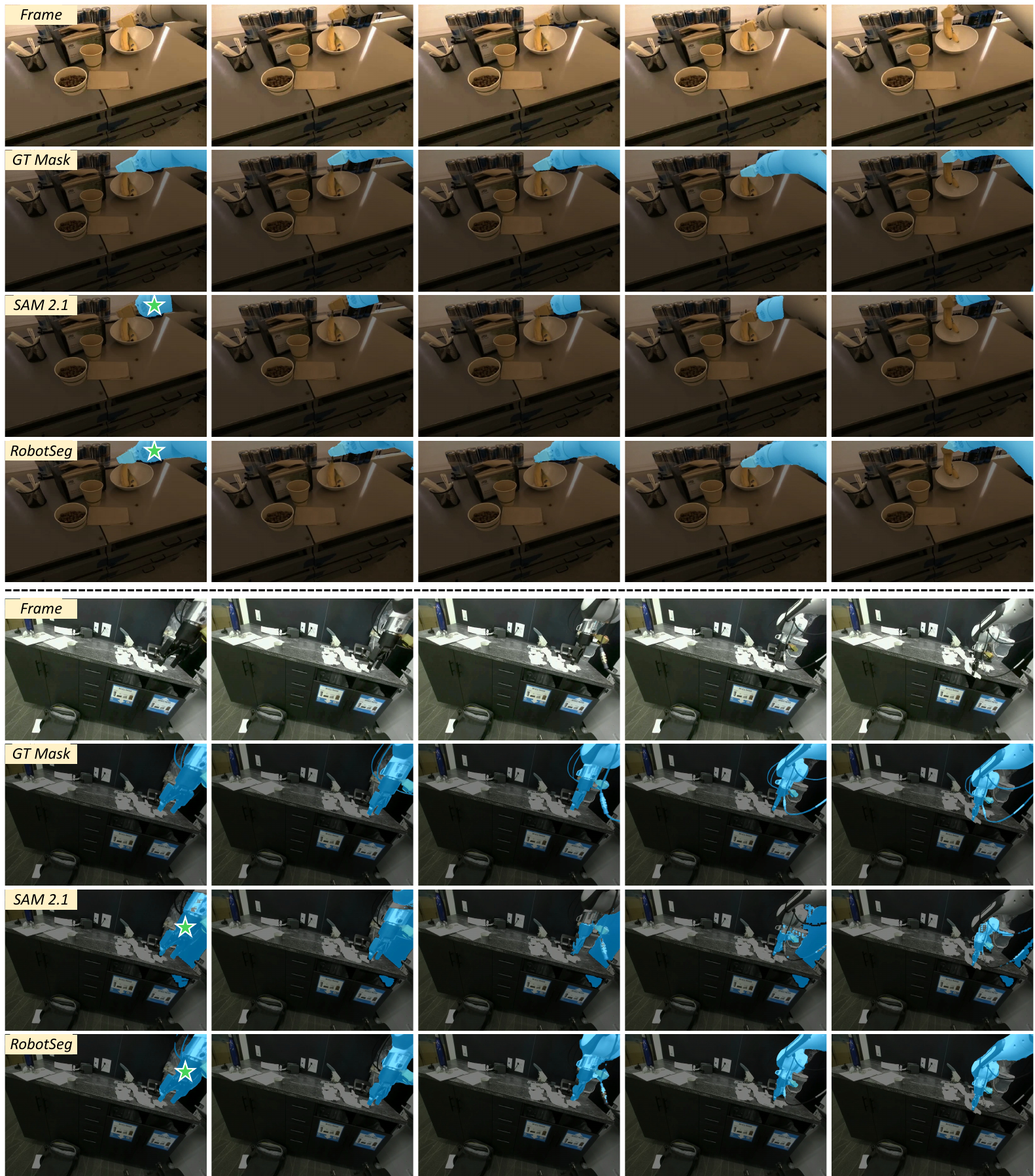}
    \caption{Comparison of robot segmentation results when a single click (green star) is given on the first video frame. SAM~2.1 \cite{sam2} often produces incomplete masks or confuses dark grippers with the background, whereas our RobotSeg generates complete and clean segmentation across frames.}
    \label{fig:visual_1c}
\end{figure*}

\begin{figure*}[t]
    \centering
    \vspace{6mm}
    \includegraphics[width=1\linewidth]{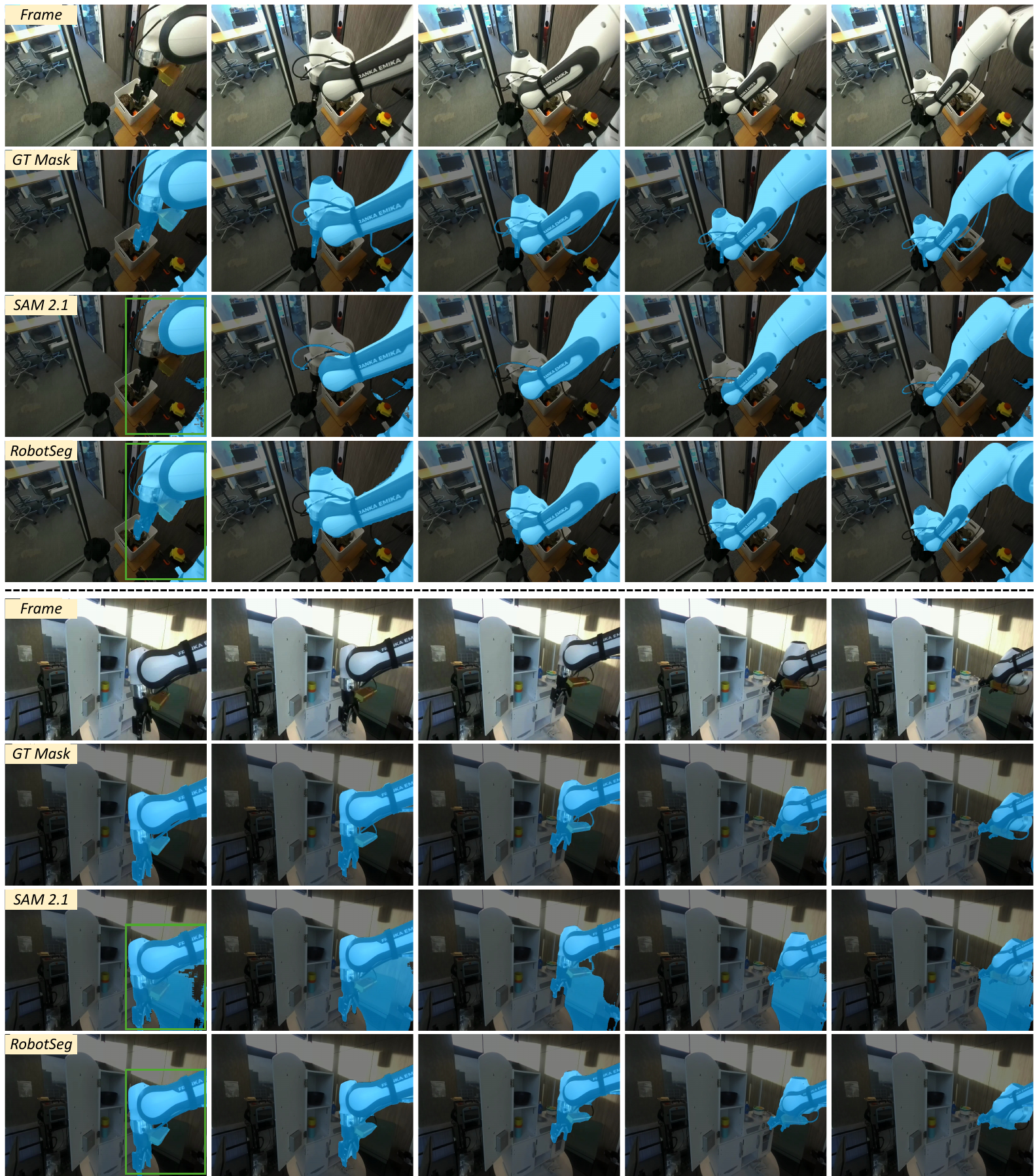}
    \caption{Comparison of robot segmentation results when a bounding box (green rectangle) is provided on the first video frame. SAM~2.1 \cite{sam2} shows inconsistent behavior, including under- and over-segmentation, while our RobotSeg maintains accurate and temporally consistent predictions.}
    \label{fig:visual_bb}
\end{figure*}

Figure \ref{fig:visual_robot}, \ref{fig:visual_arm}, and \ref{fig:visual_gripper} provide additional qualitative comparisons under the automatic segmentation setting, covering the three levels of robot granularity: whole robot, robot arm, and robot gripper. These examples complement the limited visualizations included in the main paper due to space constraints and further highlight the challenges posed by diverse embodiments and cluttered scenes. We also include prompt-based comparisons in Figure \ref{fig:visual_1c} and \ref{fig:visual_bb}, where segmentation is guided by a single click or a bounding box on the first video frame. Together, these qualitative results provide a comprehensive view of segmentation performance under both automatic and prompt-guided modes.

Under the \textbf{\textit{automatic segmentation}} setting, RoboEngine \cite{yuan2025roboengine} exhibits clear inaccuracies and temporal inconsistencies for the whole robot segmentation (Figure \ref{fig:visual_robot}). In the top example, it incorrectly segments background clothing as part of the robot in the second and fourth columns, while only loosely capturing the robot region in the third column. Similar issues appear throughout the sequence. For the robot arm segmentation (Figure \ref{fig:visual_arm}), RoboEngine cannot distinguish the arm from the gripper, so we instead compare our RobotSeg with EVF-SAM \cite{zhang2024evf}, which achieves the best performance among language-conditioned methods. In the top example, EVF-SAM identifies only the robot base while entirely missing the articulated arm. In the bottom example, it mistakenly labels a coffee machine with similar color as the robot and produces temporally unstable predictions. In the robot gripper segmentation (Figure \ref{fig:visual_gripper}), EVF-SAM again fails to localize the correct component: in the top example, it segments the arm instead of the gripper, and in the bottom example, the final two columns incorrectly segment clothing as the gripper. In contrast, our RobotSeg consistently produces accurate, component-specific masks with stable temporal behavior across diverse embodiments and complex backgrounds, demonstrating strong robustness in the automatic segmentation setting.

Figure \ref{fig:visual_1c} and \ref{fig:visual_bb} show comparisons with SAM 2.1 \cite{sam2} under \textbf{\textit{prompt-based segmentation}}. In the 1-click setting (Figure \ref{fig:visual_1c}), a single point is provided on the first video frame (green star). In the top example, SAM 2.1 only segments a partial portion of the robot, while in the bottom example, it confuses the black robot gripper with the similarly colored background, leading to incomplete or mixed masks. RobotSeg, however, generates complete and clean robot masks without confusing foreground and background. In the bounding-box setting (Figure \ref{fig:visual_bb}), where a box is given on the first video frame (green rectangle), SAM 2.1 again shows inconsistency: the top example under-segments the robot, while the bottom example over-segments into surrounding regions. In contrast, our RobotSeg delivers stable and temporally consistent results in both sequences. Overall, the prompt-based comparisons confirm that our RobotSeg remains robust and accurate when initialized with minimal user input.

Across all automatic and prompt-guided settings, these visual comparisons collectively demonstrate that RobotSeg provides accurate, temporally consistent, and embodiment-robust robot segmentation, significantly outperforming existing methods under challenging real-world scenarios.

\section{Computational Efficiency Analysis}
\label{sec:flops}
We focus in this work on developing RobotSeg, the first foundation model for robot segmentation that supports both images and videos. To provide a complete view of its practical applicability, we report computational efficiency comparisons in \autoref{tab:flops}.

\begin{table}[tbp]
\caption{Comparison of the computational efficiency of different methods. For each method, we list model parameters, FLOPs, and average inference time per frame.}
\vspace{-5pt}
\label{tab:flops}
    \centering
    \footnotesize
    \renewcommand{\arraystretch}{1.1}
    \setlength{\tabcolsep}{10pt}
    \begin{tabular}{l|c|c|c}
        \hline
        \hline
        Methods & Para. (M) & FLOPs (G) & Time (ms) \\
        \hline
        \hline
        RoVi-Aug \cite{chen2025rovi} & 638.5 & 546.1 & 87.9 \\
        \hline
        RoboEngine \cite{yuan2025roboengine} & 898.4 & 1753.4 & 250.7 \\
        \hline
        \hline
        CLIPSeg \cite{luddecke2022image} & 150.8 & 98.5 & 95.3 \\
        \hline
        LISA \cite{lai2024lisa} & 13992.9 & 61820.4 & 431.3 \\
        \hline
        EVF-SAM \cite{zhang2024evf} & 898.4 & 1753.4 & 250.7 \\
        \hline
        VideoLISA \cite{bai2024one} & 4788.3 & 36300.6 & 670.8 \\
        \hline
        \hline
        SAM 2.1 \cite{sam2} & 39.0 & 284.3 & 68.6 \\
        \hline
        \hline
        SAM 3 \cite{sam3} & 860.1 & 5045.2 & 160.1 \\
        \hline
        \hline
        \textbf{RobotSeg} (Ours) & 41.3 & 319.8 & 94.2 \\
        \hline
        \hline
    \end{tabular}
    \vspace{-2pt}
\end{table}

From \autoref{tab:flops}, RobotSeg offers a clear computational advantage over existing robot segmentation baselines. Its overall FLOPs (319.8G) are substantially lower than robot-specific methods such as RoVi-Aug~\cite{chen2025rovi} and RoboEngine~\cite{yuan2025roboengine}, as well as language-conditioned approaches including LISA~\cite{lai2024lisa}, EVF-SAM~\cite{zhang2024evf}, and VideoLISA \cite{bai2024one}, all of which rely on considerably larger backbones and therefore incur much higher computational cost. Although CLIPSeg~\cite{luddecke2022image} is lightweight in terms of FLOPs, its significantly lower segmentation performance renders it unsuitable for precise robot segmentation. Compared with SAM~2.1~\cite{sam2} (284.3G), RobotSeg introduces only a modest increase in FLOPs while providing capabilities that SAM~2.1 does not support: (i) fully automatic robot segmentation without requiring manual prompts to initialize segmentation, and (ii) substantially improved segmentation quality across both images and videos. Moreover, our RobotSeg achieves an average inference time of 94.2 ms per frame (${>}10$ FPS) (measured on an NVIDIA RTX A5000 GPU), which remains competitive among high-capacity models and supports practical deployment. Overall, these results show that RobotSeg preserves computational efficiency while offering much stronger task-specific performance, making it a practical choice for both academic research and real-world deployment.

\section{Comparison with SAM 3}
\label{sec:sam3}
SAM 3 \cite{sam3} is a concept segmentation model that supports noun-phrase prompts. Although it is designed for open-vocabulary concept segmentation rather than robot-specific perception, it can be directly applied to our task by prompting it with phrases such as ``robot'', ``robot arm'', or ``robot gripper''.

\begin{figure*}[t]
    \centering
    \includegraphics[width=.85\linewidth]{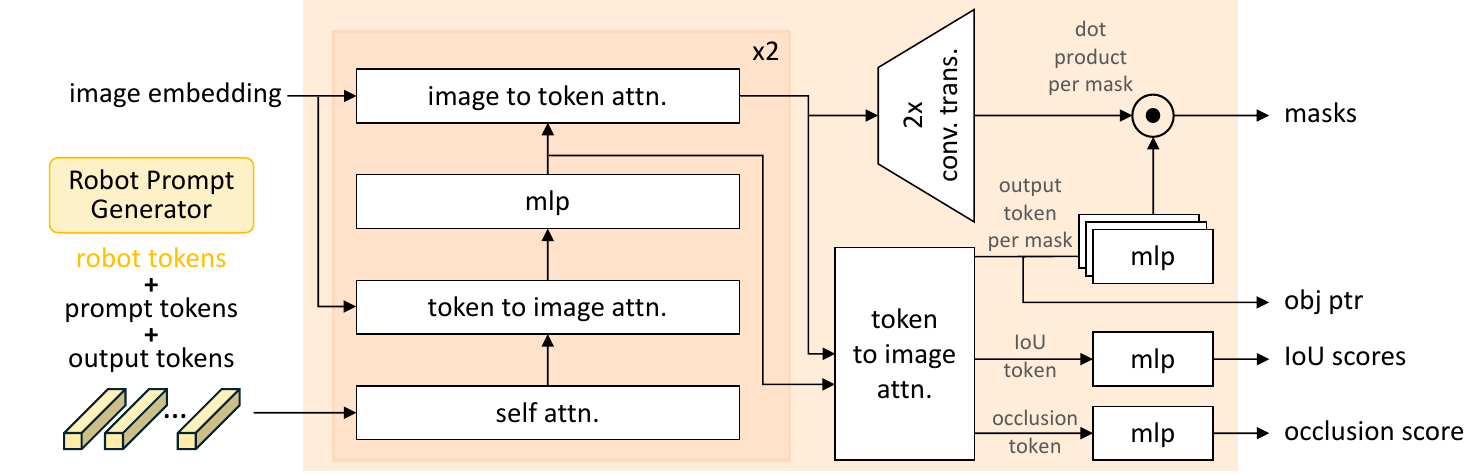}
    \caption{Architecture details of the mask decoder.}
    \label{fig:md}
    \vspace{-3mm}
\end{figure*}

\begin{figure}[t]
    \centering
    \includegraphics[width=1\linewidth]{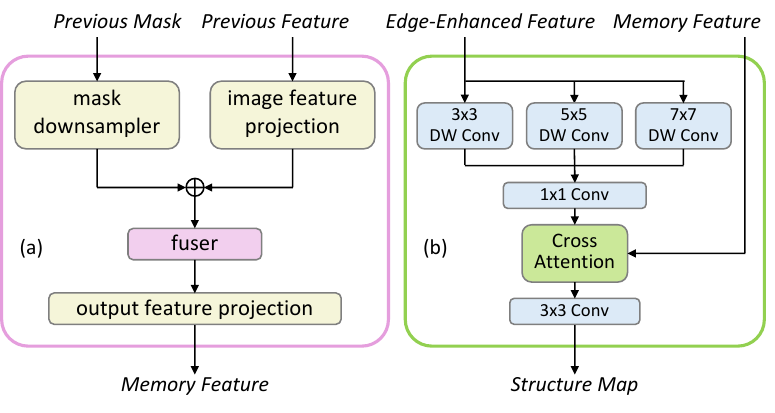}
    \vspace{-3mm}
    \caption{Architecture details of (a) the memory encoder and (b) the structure perceiver.}
    \label{fig:ab}
\end{figure}

We apply SAM 3 to our VRS benchmark using concept prompts corresponding to robot components. As summarized in Table~\ref{tab:vrs_robot},\ref{tab:vrs_arm},\ref{tab:vrs_gripper}, SAM 3 is indeed capable of segmenting robots at the concept level, but its performance remains noticeably below that of RobotSeg. SAM 3 is not tailored for articulated robot geometry, fine-grained arm–gripper separation, or temporally consistent segmentation in complex robot manipulation videos, which limits its performance on these scenarios. In contrast, RobotSeg incorporates structure-enhanced memory association and robot-specific prompt generation, achieving substantially higher accuracy and stability.

Overall, these results indicate that while SAM 3 \cite{sam3} provides a general open-vocabulary interface that can segment robots via concept prompts, dedicated modeling is required to achieve high-quality robot segmentation in realistic and dynamic environments.

\section{Architecture Details}
\label{sec:technical_details}
This section provides extended architectural details that complement the model description in the main manuscript.

\subsection{Memory Encoder}
The memory encoder is responsible for transforming the predicted masks and image encoder embeddings into a compact representation that can effectively guide subsequent frames. As illustrated in \autoref{fig:ab} (a), the memory encoder first processes the previous-frame mask through a downsampling branch and projects the previous image feature into a consistent embedding space. These two sources are then fused via lightweight convolutional operations to integrate both spatial cues and semantic context. The fused representation is subsequently passed through an output projection layer, producing the memory feature used for temporal guidance in the following frames.

\subsection{Structure Perceiver}
The structure perceiver uses the memory feature produced by the memory encoder to guide the extraction of robot-aware structural cues and the generation of the structure map. As illustrated in \autoref{fig:ab} (b), the edge-enhanced features are processed by a multi-scale depthwise convolution module (3$\times$3, 5$\times$5, and 7$\times$7) to capture edge patterns at different receptive fields. The aggregated multi-scale features are then aligned with the memory feature through a cross-attention layer, enabling the model to inject temporal robot context into the structural perceiving. Finally, a 3$\times$3 convolution predicts the structure map, which serves as structure guidance for refining following representations.

\subsection{Mask Decoder}
The mask decoder is responsible for generating segmentation masks conditioned on image embeddings and prompt tokens. Our implementation extends the SAM mask decoder by introducing additional robot prompts, enabling automatic and more consistent robot segmentation across video frames.

In \autoref{fig:md}, the prompt tokens represent optional user guidance such as clicks or bounding boxes, while the robot tokens encode robot semantic and temporal context extracted from previous frames, guiding the segmentation in the current frame. The inclusion of robot tokens enhances temporal consistency by leveraging historical robot information. The mask decoder uses a sequence of two-way transformer blocks that perform bidirectional attention between image embeddings and token representations. Specifically, the following attention mechanisms are used:
\begin{compactenum}
    \item \textbf{Self-attention}: applied to tokens to learn interactions within the token space.
    \item \textbf{Token-to-image attention}: enables token embeddings to query relevant image features.
    \item \textbf{Image-to-token attention}: aggregates token responses into the image feature representations.
\end{compactenum}

After feature fusion through the transformer blocks, the decoder outputs multiple candidate masks to handle prompt ambiguities. Following SAM~2~\cite{sam2}, we retain only the mask with the highest predicted Intersection-over-Union (IoU) score for propagation. Additionally, the decoder includes an occlusion prediction head implemented as an MLP, which predicts whether the target robot is visible in the current frame. This capability is important for accurately handling partial or full occlusions of the robot.

\begin{figure}[t]
    \centering
    \includegraphics[width=1\linewidth]{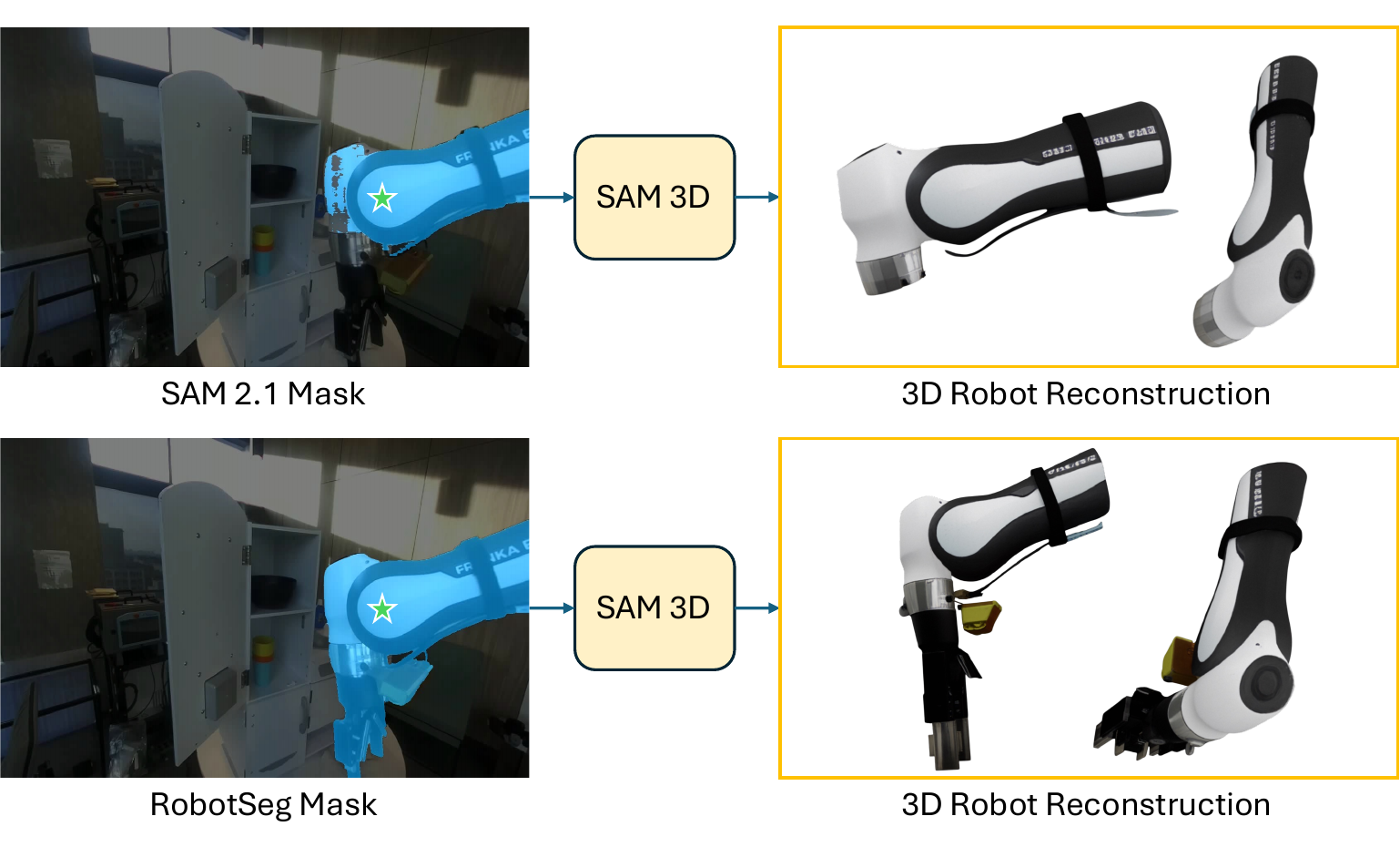}
    \vspace{-3mm}
    \caption{Compared with the incomplete robot mask from SAM 2.1 \cite{sam2} (top row), our RobotSeg provides the complete and accurate robot mask (bottom row), enabling SAM 3D \cite{sam3dobjects} to generate accurate and faithful 3D robot reconstructions.}
    \label{fig:sam3d}
    \vspace{-3mm}
\end{figure}

\section{Limitation and Future Work}
\label{sec:limitations}
While our RobotSeg demonstrates strong overall performance across diverse embodiments and challenging scenes, several limitations remain. First, although RobotSeg consistently outperforms existing models, it does not achieve the highest accuracy on every individual robot category. Certain robots with unusual appearances or scene conditions remain challenging, suggesting that further embodiment-specific modeling could provide additional gains. Second, the current design introduces moderate computational overhead compared to the original SAM~2.1. Although RobotSeg remains efficient relative to existing baselines, there is still room for reducing FLOPs and model parameters, especially for deployment on resource-constrained platforms such as mobile manipulators or embedded robotic systems.

These limitations open several promising directions for future work. One direction is to explicitly incorporate additional modalities such as depth, motion cues, or tactile signals, which may help disambiguate difficult cases where RGB appearance is insufficient. Another direction is to develop more lightweight architectures or distillation strategies that retain RobotSeg’s robustness while significantly reducing computational cost. Finally, integrating RobotSeg into closed-loop robotic systems and studying its impact on downstream tasks such as 3D robot reconstruction (\eg, \autoref{fig:sam3d}), policy learning, manipulation, and navigation represents an exciting avenue for advancing robot perception and control.